\documentclass{IEEEtran}

\usepackage{amsmath,amsfonts}
\usepackage{algorithmic}
\usepackage{algorithm}
\usepackage{array}
\usepackage{textcomp}
\usepackage{stfloats}
\usepackage{url}
\usepackage{verbatim}
\usepackage{graphicx}
\usepackage{cite}

\usepackage{hyperref}
\hypersetup{colorlinks}

\usepackage{multirow}
\usepackage{graphicx}
\usepackage{float}
\usepackage{makecell}
\usepackage{lscape}
\usepackage{algorithm} 
\usepackage{listings}
\usepackage{color}
\usepackage{booktabs}
\usepackage{graphicx}
\usepackage[lofdepth,lotdepth]{subfig}
\usepackage{caption,subcaption}

\usepackage{lipsum}

\usepackage[table,xcdraw]{xcolor}
\usepackage{CJKutf8}
\begin{document}
\begin{CJK*}{UTF8}{gbsn}

\title{RemoteCLIP: A Vision Language Foundation Model for Remote Sensing}

% \author{Fan~Liu (刘凡),~\IEEEmembership{Member,~IEEE,}
%         Delong~Chen (陈德龙),
%         Zhangqingyun~Guan (管张青云),\\
%         Xiaocong~Zhou (周晓聪),
%         Jiale~Zhu (朱佳乐),
%         Qiaolin~Ye (业巧林)~\IEEEmembership{Member,~IEEE},
%         Liyong~Fu (符利勇),
%         Jun~Zhou (周峻),~\IEEEmembership{Senior Member,~IEEE}
\author{Fan~Liu,~\IEEEmembership{Member,~IEEE,}
        Delong~Chen,
        Zhangqingyun~Guan,
        Xiaocong~Zhou,\\
        Jiale~Zhu,
        Qiaolin~Ye~\IEEEmembership{Member,~IEEE},
        Liyong~Fu,
        Jun~Zhou,~\IEEEmembership{Senior Member,~IEEE}
% \thanks{}% 
\thanks{Fan Liu and Delong Chen contributed equally. Corresponding author: Fan Liu (fanliu@hhu.edu.cn) and Delong Chen (delong.chen@connect.ust.hk).}% 
\thanks{Fan Liu, Qingyun~Guanzhang, Xiaocong~Zhou, and Jiale~Zhu are with the College of Computer and Information, Hohai University, Nanjing 210098, China.}% 
\thanks{Delong Chen is with Department of Electronic and Computer Engineering, Hong Kong University of Science and Technology, Hong Kong, China.}
\thanks{Qiaolin~Ye is with College of Information Science and Technology, Nanjing Forestry University, Nanjing 210037, China.}
\thanks{Liyong~Fu is with Institute of Forest Resource Information Techniques, Chinese Academy of Forestry, Beijing 100091, China.}
\thanks{Jun Zhou is with the School of Information and Communication Technology, Griffith University, Nathan, Queensland 4111, Australia.}%
}
% \markboth{Pre-print}%


\maketitle

\begin{abstract}
General-purpose foundation models have led to recent breakthroughs in artificial intelligence. In remote sensing, self-supervised learning (SSL) and Masked Image Modeling (MIM) have been adopted to build foundation models. However, these models primarily learn low-level features and require annotated data for fine-tuning. Moreover, they are inapplicable for retrieval and zero-shot applications due to the lack of language understanding. To address these limitations, we propose RemoteCLIP, the first vision-language foundation model for remote sensing that aims to learn robust visual features with rich semantics and aligned text embeddings for seamless downstream application. To address the scarcity of pre-training data, we leverage data scaling which converts heterogeneous annotations into a unified image-caption data format based on Box-to-Caption (B2C) and Mask-to-Box (M2B) conversion. By further incorporating UAV imagery, we produce a 12 $\times$ larger pretraining dataset than the combination of all available datasets. RemoteCLIP can be applied to a variety of downstream tasks, including zero-shot image classification, linear probing, \textit{k}-NN classification, few-shot classification, image-text retrieval, and object counting in remote sensing images. Evaluation on 16 datasets, including a newly introduced RemoteCount benchmark to test the object counting ability, shows that RemoteCLIP consistently outperforms baseline foundation models across different model scales. Impressively, RemoteCLIP beats the state-of-the-art method by 9.14\% mean recall on the RSITMD dataset and 8.92\% on the RSICD dataset. For zero-shot classification, our RemoteCLIP outperforms the CLIP baseline by up to 6.39\%  average accuracy on 12 downstream datasets.  
\end{abstract}


\section{Introduction}
\label{sec:introduction}

    \IEEEPARstart{F}{oundation} models~\cite{Bommasani2021OnTO} are becoming increasingly important in the field of Artificial Intelligence (AI). Compared to small, specialized models tailored for specific tasks or domains, ``one-for-all''-style general-purpose foundation models typically exhibit superior capabilities and generalization abilities in a wide range of downstream tasks. Numerous foundation models have emerged in recent years, such as SimCLR~\cite{Chen2020ASF}, MAE~\cite{He2021MaskedAA}, and SAM~\cite{Kirillov2023SegmentA} for computer vision, BERT~\cite{Devlin2019BERTPO} and GPT series~\cite{Brown2020LanguageMA,OpenAI2023GPT4TR} for natural language processing, also CLIP~\cite{Radford2021LearningTV} and Flamingo~\cite{Alayrac2022FlamingoAV} for vision-language. 
    
    Meanwhile, our remote sensing community is also progressing towards developing foundation models for satellite imagery analysis. To date, the prevailing approaches are primarily inspired by the success of self-supervised learning (SSL) in computer vision, particularly the Masked Image Modeling (MIM)~\cite{He2021MaskedAA,Bao2021BEiTBP,Xie2021SimMIMAS} method. Several recent studies, including SatMAE~\cite{Cong2022SatMAEPT}, Scale-MAE~\cite{Reed2022ScaleMAEAS}, ViTAE~\cite{Wang2022AdvancingPV}, Billion-scale MAE~\cite{Cha2023ABF}, RingMo~\cite{Sun2022RingMoAR}, and GFM~\cite{Mendieta2023GFMBG}, have employed MIM on large Vision Transformers (ViT) and large-scale satellite imagery datasets, yielding encouraging results.
    
    % Nevertheless, recent studies have revealed that the MIM method primarily learns low-level visual features instead of high-level semantics. For example, Kong et al.~\cite{Kong2022UnderstandingMI} proved that MIM pretraining is equivalent to learning occlusion-invariant visual features. Park et al.~\cite{Park2023WhatDS} showed that MIM methods prefer to learn high-frequency texture features instead of capturing longer-range global patterns. Although such low-level features are proven to be beneficial to visual recognition tasks in general domains (\textit{e.g.,} natural images in ImageNet)~\cite{Geirhos2018ImageNettrainedCA}, it is not clear whether they are optimal for the satellite imagery domain. 
    
    \textcolor{black}{
        Nevertheless, there are two key limitations for MIM-based remote sensing foundation models. \textit{\textbf{Firstly}}, Kong et al.~\cite{Kong2022UnderstandingMI} and Li et al.~\cite{Li2023Architecture} revealed that the MIM methods primarily learn occlusion invariant features: they implicitly align two views of the original image -- one with random mask and one with the complementary mask. Occlusion invariance is important for natural image recognition as there would be inevitable yet frequent object occlusions for views on the ground. However, the aerial view of remote sensing imagery enables unobstructed perception, making occlusion invariance much less necessary. \textit{\textbf{Secondly}}, both theoretical and empirical studies shows that MIM learns low-level features and lack semantics~\cite{Kong2023Understanding, Xie2023Revealing}. Such features have advantages for relatively low-level dense prediction tasks such as detection and segmentation, but is not optimal for high-level semantic recognition tasks, especially in the linear probing and few-shot learning setting as discussed in~\cite{Tao2023Siamese,Tukra2023Improving}. Meanwhile, Park et al.~\cite{Park2023WhatDS} showed that MIM methods prefer to learn high-frequency texture features instead of capturing longer-range global patterns, which is in stark contrast to human behavioural evidence and limits model performance and robustness~\cite{Geirhos2018ImageNettrainedCA}.
    }

    % Furthermore, all of existing foundation models require annotated data and an additional fine-tuning stage, so that they can be adapted to downstream tasks (\textit{e.g., } scenes classification). They are unable to perform \textit{zero-shot} inference like the CLIP model~\cite{Radford2021LearningTV} due to the lack of joint modeling of vision and language. As noted in recent studies~\cite{Mai2022TowardsAF,Mai2023OnTO}, multi-modality should play a crucial role in building foundation models for geospatial artificial intelligence (GeoAI). A vision-language foundation model for remote sensing could pave the way for numerous CLIP-based vision-language applications in remote sensing scenarios, such as open-vocabulary object detection, zero-shot image segmentation, text-to-image generation and editing, and multimodal large language models.
    
    \textcolor{black}{
        Furthermore, all of existing foundation models require annotated data and an additional fine-tuning stage to be adapted to downstream tasks. They are unable to perform \textit{zero-shot} inference like the CLIP model~\cite{Radford2021LearningTV} due to the lack of joint modeling and alignment of vision and language. As advocated by Mai et al. in their recent vision papers~\cite{Mai2022TowardsAF,Mai2023OnTO}, multi-modality should play a crucial role in building foundation models for geospatial artificial intelligence (GeoAI). A vision-language foundation model for remote sensing could pave the way for numerous CLIP-based vision-language applications in remote sensing scenarios, such as open-vocabulary object detection, zero-shot image segmentation, text-to-image generation and editing, and multimodal large language models (LLMs).
    }

    In this paper, we developed a \textit{vision-language} foundation model for remote sensing. Our goal is to learn robust visual features with rich semantics of satellite imaginary visual concepts while simultaneously learning text embeddings that aligned well with the visual features, which enables the learned aligned vision-language representations to be seamlessly applied into different downstream tasks and domains. To achieve this objective, the primary challenge that we faced is the scarcity of pre-training data. Although some recent works introduced high-quality human-annotated satellite imaginary captioning datasets~\cite{Yuan2021ExploringAF,Lu2017ExploringMA,Yang2010BagofvisualwordsAS}, their scale still remains much insufficient -- all existing datasets contain fewer than 10k samples, we found that training a large vision language foundation model on such dataset results in a severe over-fitting phenomenon. 
    
    To tackle this issue, we perform data scaling based on an ensemble of a wide range of remote sensing datasets, expanding the pre-training data to 12$\times$ larger than the combination of all available open datasets~\cite{Yuan2021ExploringAF,Lu2017ExploringMA,Yang2010BagofvisualwordsAS}. We convert heterogeneous annotations, including object detection bounding boxes and semantic segmentation maps, into a unified image-caption data format based on proposed mask-to-box (M2C) and box-to-caption (B2C) generation strategies. We also incorporate Unmanned aerial vehicle (UAV) imagery to further enhance the diversity of the pre-training data. We train the model to optimize the InfoNCE loss, a lower bound of mutual information between paired image and text samples, to align the vision language representations. After pre-training, we apply the resulting foundation model, which is named RemoteCLIP, to a diverse set of downstream applications, including zero-shot image classification, linear probing, \textit{k}-NN classification, few-shot classification, and image-text retrieval in remote sensing datasets. We also develop a novel benchmark, ``RemoteCount'', based on automatically-created counterfactual examples to test the object counting ability. Our comprehensive evaluation on a total of 16 datasets demonstrates that our RemoteCLIP yields superior performance compared to various baseline foundation models, which are consistent across different model scales from ResNet-50 with 38 million parameters to ViT-Large-14 with 304 million parameters. 

    % Here we briefly summarize our empirical results. We first evaluate our RemoteCLIP on cross-modal retrieval (image-to-text retrieval and text-to-image retrieval) \textcolor{black}{and achieve the new state-of-the-art (SOTA) performance on all three} remote sensing retrieval benchmarks. Specifically, RemoteCLIP \textcolor{black}{outperforms the current SOTA method}~\cite{AlRahhal2022MultilanguageTF} by 9.14\% mean recall on the RSITMD~\cite{Yuan2021ExploringAF} dataset, and by 8.92\% mean recall on the RSICD~\cite{Lu2017ExploringMA} dataset. For zero-shot classification, our RemoteCLIP outperforms the CLIP baseline by up to 6.39\% average accuracy on 12 downstream datasets. For the object counting evaluation on our newly introduced RemoteCount dataset, RemoteCLIP outperforms CLIP by 21.02\% top-1 accuracy.  \textcolor{red}{(It is not necessary to include the detailed results here. This paragraph can be removed.)}
    
    The contributions of this paper are summarized as follows:
    
    \begin{itemize}
        \item \textbf{A large-scale dataset for the remote sensing domain}: This paper introduces a comprehensive dataset that combines a wide range of remote sensing datasets. This dataset is 12 times larger than the combination of RSITMD, RSICD, and UCM datasets~\cite{Yuan2021ExploringAF,Lu2017ExploringMA,Yang2010BagofvisualwordsAS}, addressing the scarcity of pre-training data in remote sensing.
        
        \item \textbf{A novel vision-language foundation model for remote sensing} : We propose a novel vision-language foundation model called RemoteCLIP. With the large-scale pretraining dataset, this model is trained to align vision-language representations and learns robust visual features with rich semantics of satellite imagery visual concepts. We make our pretrained models available at \url{https://github.com/ChenDelong1999/RemoteCLIP}.
        
        \item \textbf{Diverse downstream applications for remote sensing}: The effectiveness of RemoteCLIP is evaluated on various downstream tasks, including cross-modal retrieval, zero-/few-/full-shot satellite imagery classification, and object counting. We also introduce a new benchmark, called RemoteCount, for object counting in remote sensing imagery.
    \end{itemize}

    The remainder of this paper is structured as follows. In Section~\ref{sec:related_work}, we review related literature on vision language models and existing foundation models in remote sensing. Section~\ref{sec:remoteclip} introduces our methodology of building RemoteCLIP -- we first prove in Section~\ref{sec:clip} that the vision-language representation of large CLIP models is very powerful for remote sensing tasks, but the data for continual pretraining is a major bottleneck to improve CLIP's performance further. Then, in Section~\ref{sec:data_scaling}, we describe the details of how we perform data scaling to address this issue. A comprehensive analysis of our new dataset is given in Section~\ref{sec:data_analysis}. Section~\ref{sec:experiments} presents our empirical evaluation of RemoteCLIP. Finally, Section~\ref{sec:conclusion} presents our discussion of the advantages and limitations of RemoteCLIP and concludes this paper.

\section{Related Work}
\label{sec:related_work}

\subsection{\textcolor{black}{Self-supervised Foundation Models for Remote Sensing}}

    \textcolor{black}{Foundation models, capable of handling multiple downstream tasks following large-scale pretraining, have recently emerged as a focal point in AI research. Concurrently, the remote sensing community is endeavoring to construct foundational models for GeoAI. The current efforts predominantly rely on Self-Supervised Learning (SSL), which devises pretext tasks to cultivate robust visual representations. This approach has made significant strides in recent years. Mainstream methods can be broadly classified into two categories: contrastive methods (also referred to as Siamese structures, Joint Embedding Predictive Architectures (JEPA), or augmentation-based methods) and generative methods. The research landscape of SSL-based foundation models for remote sensing closely mirrors this categorization~\cite{Wang2022Self}.}

    \textcolor{black}{\textbf{Contrastive Learning}: Beyond the standard data augmentation techniques employed in natural imagery, several researchers have proposed unique approaches for adapting standard SSL methods to remote sensing imagery. Kang et al.~\cite{Kang2020DeepUE}, Jung et al.~\cite{Jung2022ContrastiveSL}, and Jean et al.~\cite{Jean2018Tile2VecUR} utilized spatial neighbors as augmented data. Zhao et al.~\cite{Zhao2020WhenSL} incorporated random rotations (90$^{\circ}$, 180$^{\circ}$, 270$^{\circ}$) as augmented data. Li et al.~\cite{Li2021GeographicalKR} distilled geographical vegetation for the same purpose. Stojnic et al.~\cite{Stojnic2021Self} applied Contrastive Multiview Coding (CMC) to learn schematic representations, which were subsequently adapted for downstream classification tasks. Xiao et al.~\cite{Xiao2023degrade} demonstrated that contrastive learning can enhance the super-resolution task in remote sensing.}

    \textcolor{black}{\textbf{Generative Learning}: Masked Image Modelling (MIM)-based models are widely recognized as the leading methods for generative modeling. Several remote sensing models, grounded in the MIM methodology, primarily aim to incorporate new properties into the standard MIM framework. These include scale-invariance (Scale-MAE~\cite{Reed2022ScaleMAEAS}), temporal information (SatMAE~\cite{Cong2022SatMAEPT}), and temporal invariance (SeCo~\cite{Maas2021SeasonalCU}), among others. A recent trend in research has been to scale up the MIM model, with examples such as RingMo~\cite{Sun2022RingMoAR}, billion-scale MAE~\cite{Cha2023ABF}, and VITAE~\cite{Wang2022AdvancingPV}. These have garnered considerable attention.}

\subsection{\textcolor{black}{Vision Language Models for Remote Sensing}}
    \textcolor{black}{The integration of image and text content has been a longstanding challenge and a focal point of research in the field of artificial intelligence~\cite{Du2022ASO, Long2022VisionandLanguagePM, Gan2022VisionLanguagePB, Zhang2023VisionLanguageMF}. This challenge is particularly significant in the realm of remote sensing, where the interpretation of complex satellite imagery and associated semantic meaning is crucial. }

    \textcolor{black}{\textbf{Image-text Retrieval Models for Remote Sensing}: The initial efforts in remote sensing retrieval were spearheaded by Abdullah et al.~\cite{Ali2020TextRSDB} and Rahhal et al.~\cite{AlRahhal2020DeepUE}, who employed CNN to encode images and LSTM to encode text captions. To endow the model with the ability to comprehend satellite images on a global and local scale, Yuan et al.~\cite{Yuan2022RemoteSC} introduced a novel framework that utilizes a dynamic fusion module. Rahhal et al.~\cite{AlRahhal2022MultilanguageTF} proposed a multi-language framework, comprising a language encoder, to adapt to the remote sensing semantics of various languages. Subsequently, a growing body of research, including CMFM-Net~\cite{Yu2023Text}, HyperMatch~\cite{Yao2023Hypergraph}, KCR~\cite{Mi2022Knowledge}, HVSA~\cite{Zhang2023Hypersphere}, and others, has harnessed the process of image-text retrieval for knowledge acquisition. However, the efficacy of these vision language models in downstream applications beyond retrieval remains unverified.}
    
    \textcolor{black}{\textbf{CLIP-based Models}: In the seminal work of CLIP~\cite{Radford2021LearningTV,Jia2021ScalingUV}, a two-tower model was trained to contrastively align the representations of a vast number of image-text pairs sourced from the Internet. Recent advancements in CLIP models have primarily concentrated on scaling the model size and data size~\cite{Cherti2022ReproducibleSL}, incorporating self-supervision~\cite{Mu2021SLIPSM, Li2021SupervisionEE, Yu2022CoCaCC}, enhancing pre-training efficiency~\cite{Li2022ScalingLP,Chen2022PrototypicalCL}, and few-shot adaptation~\cite{Zhou2021LearningTP,liu2023few}, among others. As CLIP models are trained on natural imagery, a line of research aims to develop domain-specific CLIP models. For instance, in the medical domain, ConVIRT~\cite{Zhang2020ContrastiveLO}, PubMedCLIP~\cite{Eslami2021DoesCB}, MedCLIP~\cite{Wang2022MedCLIPCL}, and BioMedCLIP~\cite{Zhang2023LargeScaleDP} have shown promising results. In another example, specialized CLIP models, based on large-scale E-commerce image-text datasets, significantly outperform the naive CLIP baseline~\cite{Dong2021M5ProductSC,Liu2023MEP3MAL,Shin2022eCLIPLV}. However, despite the concurrent work by Zhang et al.~\cite{rs5m}, which involves gathering aerial view images from large image-text datasets to train CLIP models, the exploration of CLIP models in the remote sensing area remains relatively limited.}

\section{RemoteCLIP}
\label{sec:remoteclip}

\subsection{Contrastive Language Image Pretraining}
\label{sec:clip}
Vision language models trained with the Contrastive Language Image Pretraining (CLIP)~\cite{Radford2021LearningTV} strategy have demonstrated impressive generalization ability in various vision-language learning tasks. These models, usually referred to as CLIP models, learn to group and align the representations of semantically similar samples together under cross-modal supervision mined from billion-scale image-text pairs. The CLIP model optimizes a simple InfoNCE loss function, which encourages the alignment of paired image-text samples and pushes apart mismatched samples.

Formally, CLIP is trained with a large-scale image-text dataset $\mathcal{D}=\{(x^I_i, x^T_i)\}_{i=1}^M$ that consists of a total of $M$ training samples. The goal is to learn an image encoder $f^I$ and a text encoder $f^T$ that respectively encode image sample $x^I_i$ and text sample $x^T_i$ to their latent representations, \textit{i.e.}, $f^I(x^I_i) = z^I_i\in\mathbb{R}^{d_z \times 1}$ and  $f^T(x^T_i) = z^T_i\in\mathbb{R}^{d_z \times 1}$. During pretraining, CLIP creates an instance discrimination task within each batch and optimizes the following bi-directional InfoNCE  objective, where $N$ is the batch size and $\tau$ is a learnable temperature parameter:
    % \begin{equation}\label{eq:loss_CLIP}
    % \begin{aligned}\tiny
    %     \mathcal{L}_{\text{InfoNCE}}
    %     %&= (\mathcal{L}^{I}_{\text{InfoNCE}} + \mathcal{L}^{T}_{\text{InfoNCE}})/2 \\&
    %     = - (
    %     \underbrace{
    %     \frac{1}{N} \sum_{i=1}^{N} \log \frac{\text{exp}(z^I_i \cdot z^T_i/\tau)}{\sum_{j=1}^{N}\text{exp}(z^I_i \cdot z^T_j/\tau)}}_{\text{image to text}}
    %     + \underbrace{\frac{1}{N} \sum_{i=1}^{N} \log \frac{\text{exp}(z^T_i \cdot z^I_i/\tau)}{\sum_{j=1}^{N}\text{exp}(z^T_i \cdot z^I_j/\tau)}}_{\text{text to image}})/2,
    % \end{aligned}
    % \end{equation}
    \begin{equation}\label{eq:loss_CLIP}
    \begin{aligned}
        \mathcal{L}_{\text{InfoNCE}}
        % = &(\mathcal{L}^{I}_{\text{InfoNCE}} + \mathcal{L}^{T}_{\text{InfoNCE}})/2 \\
        = &- (
        \underbrace{
        \frac{1}{N} \sum_{i=1}^{N} \log \frac{\text{exp}(z^I_i \cdot z^T_i/\tau_{\text{CLIP}})}{\sum_{j=1}^{N}\text{exp}(z^I_i \cdot z^T_j/\tau_{\text{CLIP}})}}_{\text{image to text}} \\
        &+ \underbrace{\frac{1}{N} \sum_{i=1}^{N} \log \frac{\text{exp}(z^T_i \cdot z^I_i/\tau_{\text{CLIP}})}{\sum_{j=1}^{N}\text{exp}(z^T_i \cdot z^I_j/\tau_{\text{CLIP}})}}_{\text{text to image}})/2,
    \end{aligned}
    \end{equation}
   
According to Chen et al.~\cite{Chen2022PrototypicalCL}, optimizing $\mathcal{L}_{\text{InfoNCE}}$ brings the following two important properties to the CLIP model:

\begin{itemize}
    \item     Representation alignment: it produces high similarity $z^I_i \cdot z^T_i$  of paired image and text samples $x^I_i, x^T_i$, and low similarity $z^I_i \cdot z^T_j (i \neq j)$ between the unpaired samples $x^I_i, x^T_j$. Generally, perfect representation alignment yields strong downstream performance on cross-modal retrieval tasks.
    \item             Representation grouping: it means that (uni-modal) representations of semantically similar samples are grouped together, while those of dissimilar samples should be pulled apart. Perfect representation grouping yields strong uni-modal recognition (\textit{e.g.,} linear classification) performance. 
\end{itemize}

      While fulfilling perfect representation alignment and representation grouping simultaneously, coupled with a large dataset containing sufficient open-set concepts, the model can achieve strong zero-shot classification performance.

\subsubsection{Large CLIP is also a strong model for remote sensing tasks}

CLIP models do not have any special designs to optimize their performance in the remote sensing domain, and they have shown diverse zero-shot performance on remote sensing benchmarks. In the original CLIP paper, OpenAI researchers evaluated CLIP's scene recognition performance on zero-shot benchmarks EuroSAT and RESISC45. The performance of the largest CLIP (ViT-Large-14-336) is only 59.6\% and 71.7\% respectively. The recent study on Satellite ImageNet (SATIN) dataset~\cite{Roberts2023SATINAM} also confirmed that the zero-shot performance of the CLIP family is unsatisfactory. However, the linear probing accuracy of CLIP reaches 98.1\% and 94.9\% on EuroSAT and RESISC45 in OpenAI's evaluation, outperforming all other 11 compared foundation visual models including both fully-supervised and self-supervised models. It shows that large-scale contrastive image text pretraining produces high-quality visual representations that are suitable for the remote sensing domain, but at the same time, the cross-modal alignment property of such representations is unsatisfactory.

To have a more thorough understanding of the potential of CLIP models for remote sensing vision language tasks, we perform a comprehensive evaluation of CLIP's zero-shot retrieval on three commonly used remote sensing retrieval datasets: RSITMD, RSICD, and UCM (which we denote as \texttt{RET-3}). We use the pretrained weights provided by both OpenAI and OpenCLIP, covering models from ResNet-50 (38M parameters) to ViT-G-14 (1.8B parameters). We also compar representative single-tower vision language models including ALBEF and BLIP. 

\begin{figure}
    \centering
    \includegraphics[width=\linewidth]{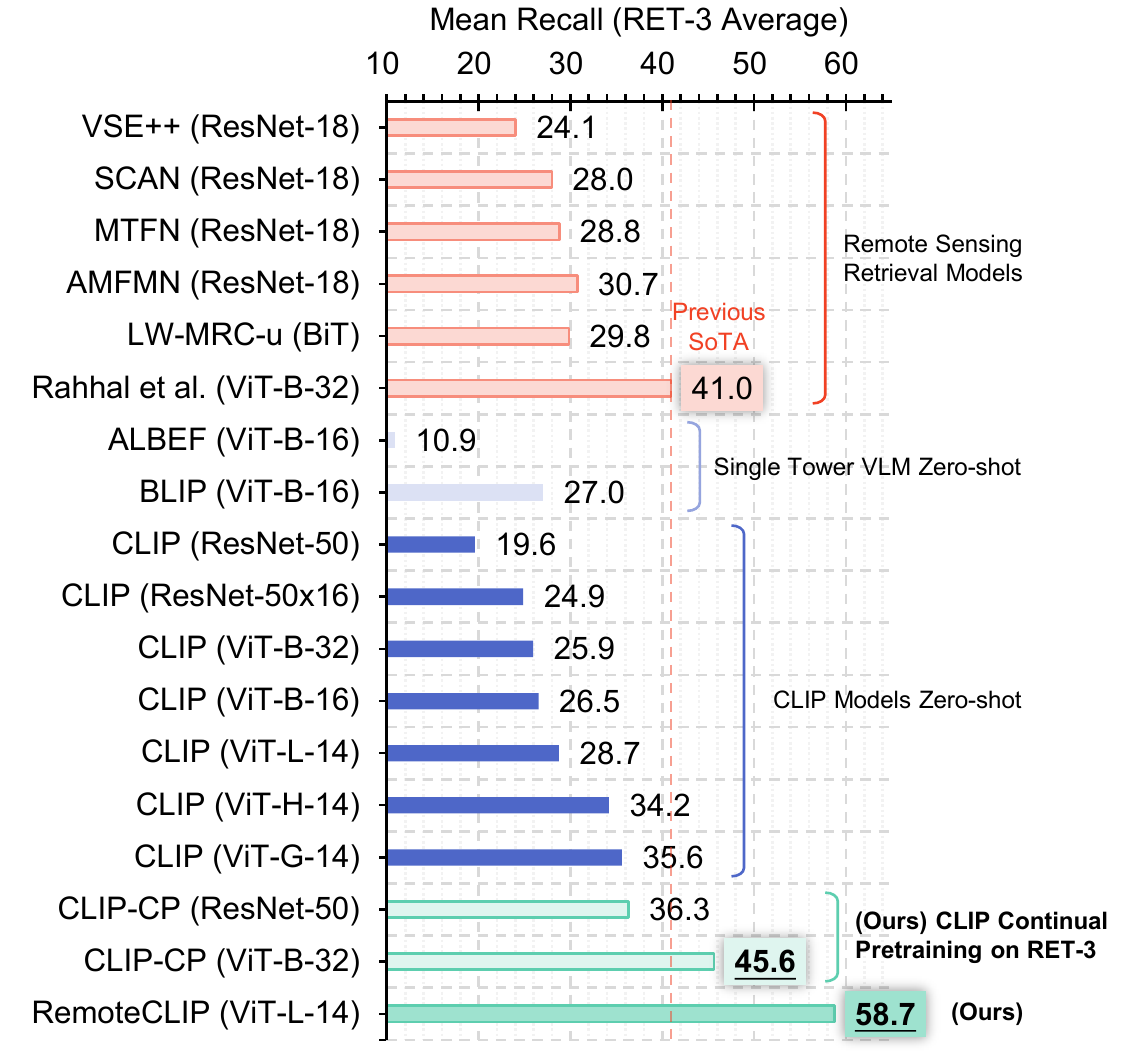}
    \caption{Averaged mean recall on three remote sensing image-text retrieval benchmarks: RSITMD, RSICD, and UCM (\texttt{RET-3}). Key findings: (1) Zero-shot retrieval of large CLIP models (\textit{e.g.,} ViT-G-14) outperforms all previous models specifically designed for remote sensing retrieval, except for the method from Rahhal \textit{et al.}~\cite{AlRahhal2022MultilanguageTF} that fine-tuned a CLIP model. (2) Simply performing continual pretraining (CLIP-CP) significantly boosts the performance of CLIP models and establishes a new SOTA model.}
    \label{fig:clip_continues_pretraining}
\end{figure}

The evaluation results are reported in Fig.~\ref{fig:clip_continues_pretraining}. We find that model size is an important factor. Larger models consistently yield better performance than smaller ones, and the largest CLIP model ViT-G-14 even surpasses all the previous retrieval methods that are specially designed for the remote sensing domain, except for the model from Rahhal \textit{et al.}~\cite{AlRahhal2022MultilanguageTF} which is based on fine-tuning of the CLIP model. In addition, large CLIP models outperform single-tower models (ALBEF and BLIP) by a large margin, demonstrating the power of simplicity - combining large-scale model and large-scale pretraining data clearly outperforms complex network structures or employing multiple loss functions.

\subsubsection{Continual pretraining on small CLIP further improves the performance}
\label{sec:clip_cl}

Given the strong results of large CLIP models on remote sensing tasks, a natural question is whether we can further improve their performance using in-domain aerial imaginary data. Continual pretraining is a popular methodology to achieve this goal, which has already shown its advantages for adapting CLIP models into the medical domain~\cite{Zhang2023LargeScaleDP}. As an initial experiment, we perform continuous pretraining of the CLIP model (ResNet-50 and ViT-Base-32) on the union of three existing retrieval datasets RSITMD, RSICD, and UCM  (\texttt{RET-3}) \footnote{Similar works have been done previously on remote sensing image-text retrieval method~\cite{AlRahhal2022MultilanguageTF}, where the authors fine-tuned CLIP models (ViT-Base-32) separately on RSITMD, RSICD, and UCM and achieved SOTA performance. However, our goal is different -- we aim to build foundation vision language models based on the powerful pretrained CLIP models.}. \textcolor{black}{We used the standard CLIP training setup, following common practice~\cite{Eslami2021DoesCB,Wang2022MedCLIPCL,Zhang2023LargeScaleDP}.} The resulting models, which we denote as CLIP-CP, yield extremely powerful performance. As shown in Fig.~\ref{fig:clip_continues_pretraining}, it not only outperforms the zero-shot results of the largest CLIP model (ViT-G-14) with only 2\% (38M vs. 1.8B) parameters but also establishes a new state-of-the-art performance on these three retrieval benchmarks.

It is also clear that tuning the foundation model on a collection of datasets is beneficial. Compared to the model of Rahhal \textit{et al.}~\cite{AlRahhal2022MultilanguageTF} using the same ViT-Base-32 architecture, our approach -- continual pretraining on the \texttt{RET-3} collection -- improved the performance by a clear margin (4.6\%). This multi-dataset tuning shares a similar spirit with recent studies on vision language learning, such as InstructBLIP~\cite{Dai2023InstructBLIPTG}, PaLI-X~\cite{Chen2023PaLIXOS}, and Clever Flamingo~\cite{chen2023visual}.

Such a simple continuous pretraining strategy yields encouraging results, but it is still far from perfect:  when we try to scale up the model size (\textit{e.g.,} to ViT-Large-14), a severe over-fitting phenomenon appears. The reason is quite clear -- the dataset  used for continuous retraining is too small for a large CLIP model. The combination of all existing image-text data (\texttt{RET-3}) only has 13k samples, while the pretraining data for CLIP models usually range from several hundred million to several billion samples. This motivates us to perform data scaling to match the model capacity and complexity of large CLIP models. As shown in the last row in Fig.~\ref{fig:clip_continues_pretraining}, such data scaling yields impressive results (+17.7\% compared to the previous SOTA results). The details of our data scaling method are presented in the following section.

\subsection{Data Scaling via Annotation Unification}
\label{sec:data_scaling}

We have already shown that the vanilla CLIP model and its continual-pretrained version are promising for vision language tasks in the remote sensing domain. We have also identified that data scale is the major bottleneck limiting performance. These observations motivate us to scale up the dataset for continuous pretraining beyond the currently available image-text pairs (\texttt{RET-3} with only 13k samples). A straightforward methodology is to annotate more captions based on crowd-sourcing, but despite this being very expensive therefore significantly lowering the scalability, the annotation quality and diversity are also hard to guarantee.

\begin{figure*}
    \centering
    \includegraphics[width=\linewidth]{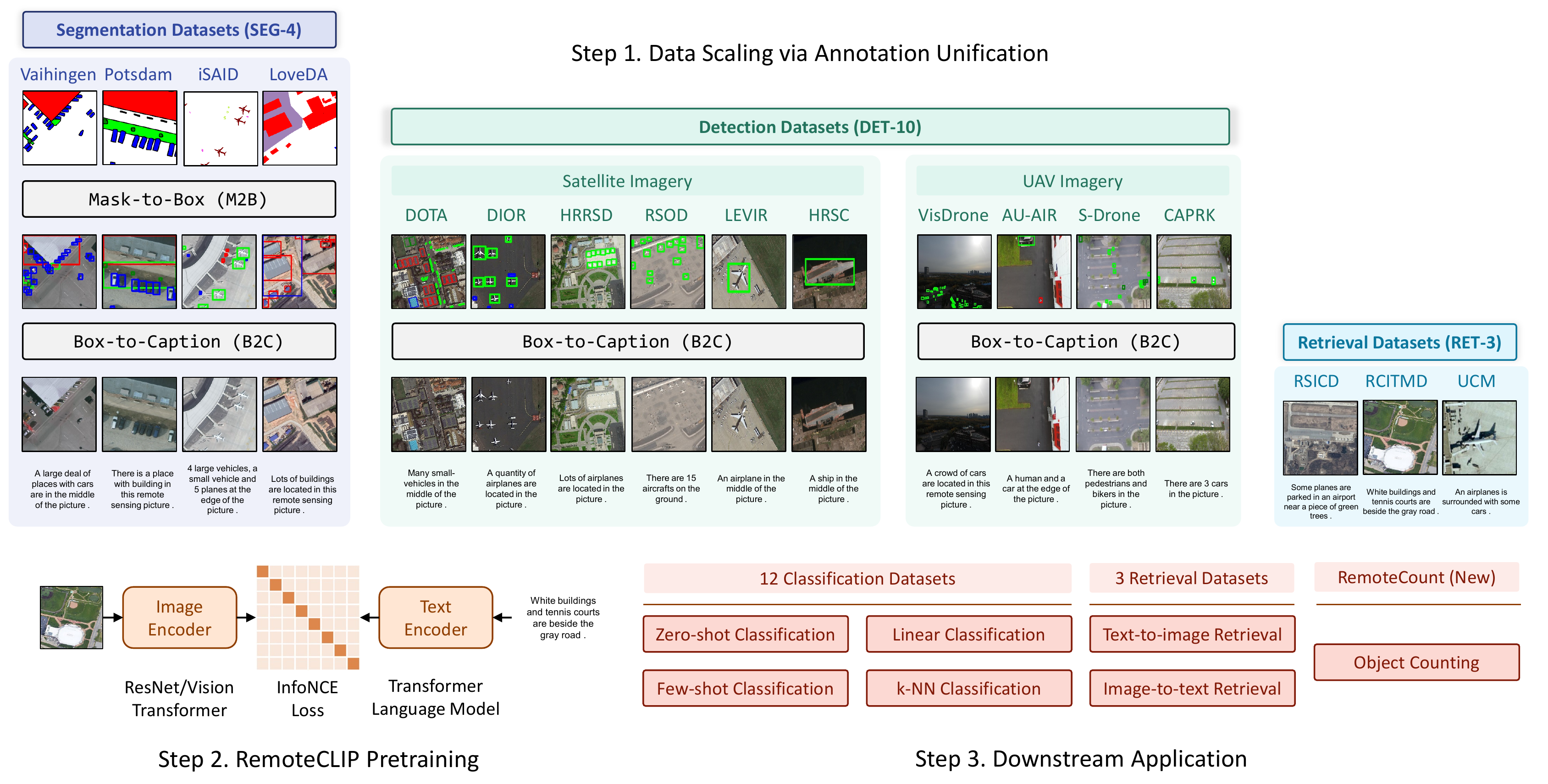}
    \caption{Overview of the RemoteCLIP pipeline. \textbf{Step 1}: RemoteCLIP is trained on a diverse collection of remote sensing datasets, covering 10 object detection datasets (\texttt{DET-10}, 6 of them are satellite imaginary datasets and 4 of them are UAV datasets), 4 remote sensing semantic segmentation datasets (\texttt{SEG-4}), and three remote sensing image-text datasets. We propose Box-to-Caption (B2C) generation and Mask-to-Box (M2B) conversion to fully utilize heterogeneous annotations, and scale up the training data to 12$\times$ of the combination of all involved image-text data. \textbf{Step 2}: We perform continual pretraining based on the CLIP model, specializing it in the remote sensing domain. \textbf{Step 3}: we perform a comprehensive evaluation on 7 tasks using 16 downstream datasets, including a newly created RemoteCount dataset, to demonstrate the strong capability and generalization ability of RemoteCLIP.}
    \label{fig:remoteclip_pipeline}
\end{figure*}

To solve this issue and thereby unleash the full potential of CLIP models, here we propose to scale up the dataset via annotation unification. We find that existing datasets annotated with object bounding boxes and class names, which were originally constructed for training object detectors, provide valuable information about the semantics within each satellite image. However, such object bounding box annotation can not be directly understood by the text encoder of CLIP, as it has only been trained on natural language captions. Therefore, we propose a Box-to-Caption (B2C) generation approach to transfer the bounding box annotations into a set of natural language captions to mitigate this gap. Further, to utilize the annotation in segmentation datasets, we propose to use a Mask-to-Box (M2B) conversion method to unify segmentation datasets into bounding box annotations, and subsequently transfer them into captioning datasets. An overview of this process is shown in Fig.~\ref{fig:remoteclip_pipeline}.

\subsubsection{Box-to-Caption (B2C) Generation}

The Box-to-Caption (B2C) generation enables the generation of textual descriptions for object detection datasets based on bounding box annotations and labels. This method employs a rule-based approach to generate five\footnote{Most image captioning/retrieval datasets such as MS-COCO, Flicker-30k, as well as three datasets in the \texttt{RET-3} collection contain five captions for each image. We choose to generate five captions to be in line with these datasets.} distinct captions that describe the objects in the image.
% , and Algorithm~\ref{alg:code} presents an overview of the B2C approach\textcolor{red}{(The algorithm does not seem to be necessary given the description below. If you have to include it, change from the pseudo-code format to the algorithm format to make it look more academic)}. 

Specifically, the first two captions are generated according to the target location (the center point of the bounding box): the first caption describes the objects in the center of the image, while the second one describes the objects that are not located in the center. This differentiation provides additional context and information about the spatial distribution of objects within the image. 

The remaining three captions are generated by considering the number of different object categories in the image. Random objects from the list of bounding box annotations are selected, and a caption is generated accordingly. In cases where the number of appearances of an object exceeds ten, a more general term (\textit{e.g., } ``many'', ``a lot of'') is used instead of the exact number to enhance the readability and variability of the captions.

\subsubsection{Mask-to-Box (M2B) Conversion}

The conversion of segmentation annotations to bounding box annotations is a crucial step for the seamless integration of segmentation datasets into the B2C generation pipeline. To perform such conversion, the segmentation mask is processed by category, encoding each pixel label corresponding to the target class. Next, the contour points of the connected regions for each class in the mask image are identified. These contour points provide the necessary information to determine the bounding box coordinates. By sorting the horizontal and vertical coordinates of the contour points, we can extract the minimum and maximum values, denoted as $(x_{min}$, $y_{min})$ and $(x_{max}$, $y_{max})$, respectively. These coordinate positions define the bounding box. 

\begin{figure}
    \centering
    \includegraphics[width=\linewidth]{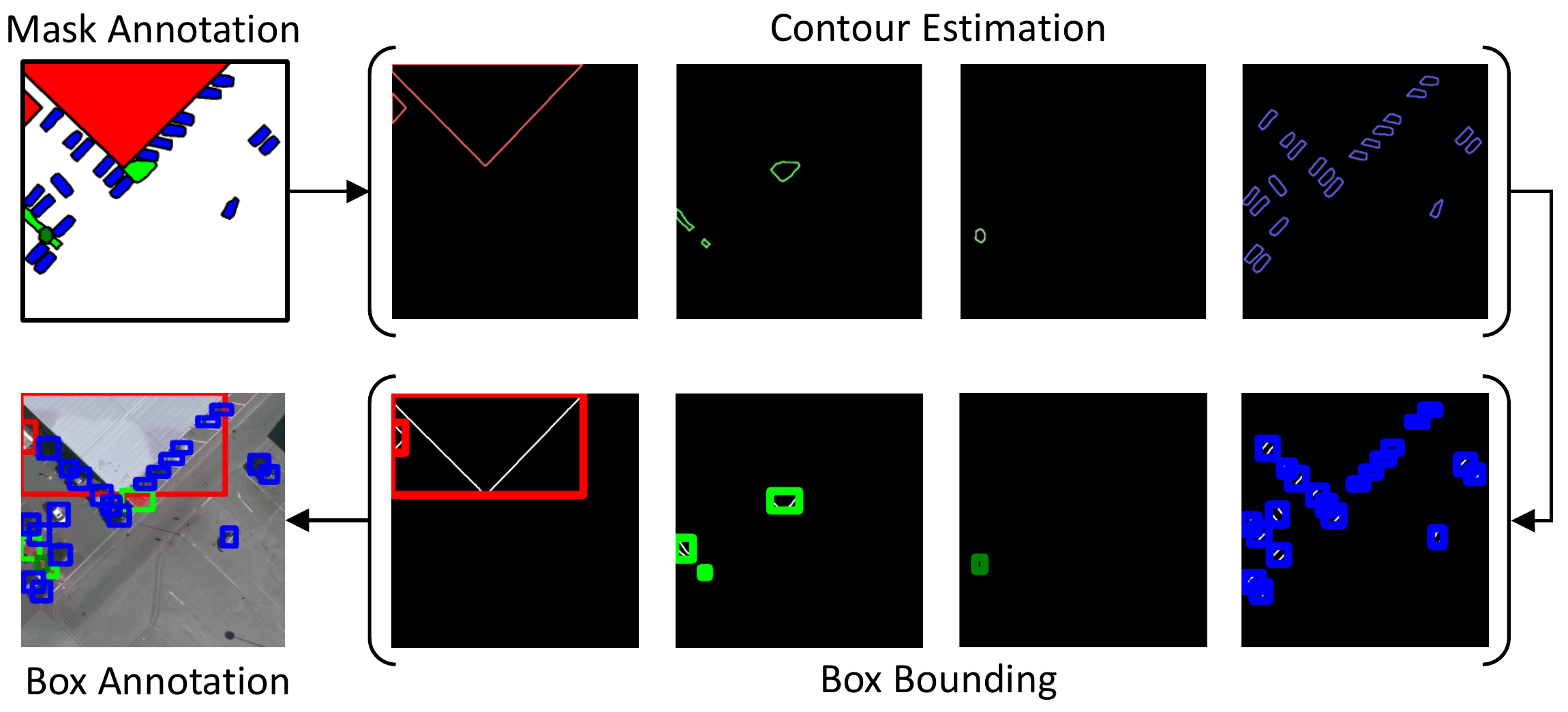}
    \caption{Mask-to-Box (M2B) implementation details. First, we get contours of per class from the input mask. Then, we select the lower left and upper right points of each contour as its bbx coordinates. Finally, we can get the bounding boxes of each category in the input mask. }
    \label{fig:M2B}
\end{figure}

\textcolor{black}{The steps mentioned above can be found in Fig.~\ref{fig:M2B}. To enhance clarity, different colors are assigned to represent bounding boxes corresponding to distinct categories. Specifically, we utilize Suzuki’s border following algorithm ~\cite{Suzuki1985Topological} for contour extraction. This algorithm defines the outer boundary and the hole boundary, and it scans the binary image from left to right to find the starting point of the outer boundary or hole boundary. By traversing the neighborhood of this starting point, the algorithm determines whether to update the pixel values based on certain rules, and ultimately extracts the hierarchical relationships among the contours. As this algorithm only supports binary images as input, we need to process segmentation masks by category. The category in need of contour extraction is identified as the foreground, while the remaining categories are treated as background.}

\textcolor{black}{Subsequently, we apply the border following algorithm to extract the topological structure of connected components within the binary mask. By sorting the contour points of the outer boundary for each connected component, the minimum and maximum values on the horizontal and vertical axes are considered as the coordinates for the horizontal bounding box of each connected component. As shown in Fig.~\ref{fig:remoteclip_pipeline}, all the semantic segmentation annotations are converted to bounding box annotations via M2B, then B2C is performed to obtain the corresponding captions}.

\subsubsection{Sample De-duplication}

RemoteCLIP is trained on a combination of datasets from different sources, and tested on a variety of downstream benchmarks, so it is essential to avoid possible test-set contamination. \textcolor{black}{P-Hash~\cite{zauner2010implementation} serves as a method used for image retrieval and similarity calculation by representing image features through  converting the image into a fixed-length hash value. We employ p-Hash-based block-wise local detection to identify duplicate images. Specifically, we generate p-Hash values for all images and partition each value into N segments. Simultaneously, N dictionaries are established, where each dictionary's key corresponds to the segment index, and the value comprises p-Hash values of all images in that segment. By traversing all dictionaries, we compute the Hamming distance between p-hash values of pair-wise images. If the distance between two images is less than the threshold of 2, they are considered duplicates. Upon observing the removed duplicated samples, when the threshold is set greater than 2, it is prone to excessive de-duplication. Conversely, de-duplication would be insufficient.} Finally, the number of removed duplicated samples ranges from 40 to 3k in different datasets.

\subsection{Data Analysis}
\label{sec:data_analysis}
Based on the proposed B2C generation and M2B conversion method, we can efficiently translate heterogeneous annotations in various detection or segmentation datasets into image-text samples based on the pipeline shown in Fig.~\ref{fig:remoteclip_pipeline}. For a deeper understanding of the dataset produced by such a scaling pipeline, in this section, we present a detailed and comprehensive analysis of this dataset. Firstly, in Table~\ref{tab:dataset_detail}, we provide the details of each source dataset used to expand the data, which can be divided into the following three groups:
% \textcolor{red}{(the Descriptions in the table can be shortened by keeping only key phrases instead of writing the full sentence. This will make the table more readable.)}
\begin{enumerate}
    \item Retrieval Data (\texttt{RET-3}). Three major image-text datasets for remote sensing, i.e., RSICD~\cite{Lu2017ExploringMA}, RSITMD~\cite{Yuan2021ExploringAF}, UCM~\cite{Yang2010BagofvisualwordsAS}, are directly adopted. Captions of these datasets are annotated by humans, which results in high caption quality but a small dataset size.
    \item Detection Data (\texttt{DET-10}). The detection dataset is the major source for dataset expansion. We combined six remote sensing datasets with object detection annotation, including DOTA~\cite{Xia2017DOTAAL}, DIOR~\cite{Li2019ObjectDI}, HRRSD~\cite{Zhang2019HierarchicalAR}, RSOD~\cite{Sun2021RSODRS}, LEVIR~\cite{Chen2020ASA} and HRSC~\cite{Liu2017AHR}. As shown in Table~\ref{tab:dataset_detail}, these datasets have a significantly higher resolution than \texttt{RET-3} datasets (at least 800$\times$600 vs. 224$\times$224). This group of datasets also exhibits high diversity as it consists of both satellite imagery and UAV imagery. The average number of objects in each image ranges from 1 (HRSC) to 70 (DOTA).
    \item Segmentation Data (\texttt{SEG-4}). Four popular remote sensing semantic segmentation datasets, including Vaihingen~\cite{vaihingen}, Postdam~\cite{potsdam}, iSAID~\cite{Zamir2019iSAIDAL}, and LoveDA~\cite{Wang2021LoveDAAR}, are adopted and translated via M2B then B2C. These datasets also have high image resolution and domain diversity. Average number of objects ranges from 2 (Vaihingen) to 33 (ISAID).
\end{enumerate}

\begin{table*}[]
\centering
\caption{dataset statics}
\label{tab:dataset_detail}
\resizebox{\textwidth}{!}{%
\begin{tabular}{cccccccc}
\toprule
\multicolumn{1}{l}{} & \textbf{Dataset} & \textbf{Year} & \textbf{\#Image} & \textbf{\#Class} & \textbf{\#Box} & \textbf{Avg. Res.} & \textbf{Description} \\
% \cline{2-8}
\midrule
\multirow{3}{*}{\texttt{RET-3}} & RSICD~\cite{Lu2017ExploringMA} & 2017 & 8483 & - & - & 224×224 & \begin{tabular}[c]{@{}c@{}}RSICD dataset contains more than ten thousands remote sensing images\end{tabular} \\
 & RSITMD~\cite{Yuan2021ExploringAF} & 2021 & 3603 & - & - & 256×256 & \begin{tabular}[c]{@{}c@{}}RSITMD dataset contains multi-source remote sensing images and textual descriptions\end{tabular} \\
 & UCMerced~\cite{Yang2010BagofvisualwordsAS} & 2018 & 1676 & - & - & 256×256 & \begin{tabular}[c]{@{}c@{}}UCMerced dataset covers 21 different scene classes, with 100 images per class.\end{tabular} \\
 \midrule
\multirow{10}{*}{\texttt{DET-10}} & AUAIR~\cite{Vujasinovic2020IntegrationOT} & 2020 & 32,823 & 8 & 132,031 & 1920×1080 & \begin{tabular}[c]{@{}c@{}}AU-AIR dataset features multi-modal sensor data, including visual, temporal, location, altitude, IMU, velocity, and more.\end{tabular} \\
 & CARPK~\cite{Hsieh2017DroneBasedOC} & 2017 & 1,568 & 1 & 106,690 & 1280×720 & \begin{tabular}[c]{@{}c@{}}CARPK dataset contains nearly 90,000 cars collected from four different parking lots by drones.\end{tabular} \\
 & DIOR~\cite{Li2019ObjectDI} & 2019 & 23,463 & 20 & 192,472 & 800×800 & DIOR dataset consists of 190,288 instances of 20 different object classes, with approximately 1,200 images per class. \\
 & DOTA~\cite{Xia2017DOTAAL} & 2017 & 1,409 & 15 & 98,990 & 1504×1395 & DOTA dataset consists of 188,282 instances of 15 different object classes, including airplanes, ships, and others. \\
 & HRRSD~\cite{Zhang2019HierarchicalAR} & 2019 & 21,761 & 13 & 57,137 & 1406×1264 & HRRSD dataset is used for studying object detection in high-resolution remote sensing images. \\
 & HRSC~\cite{Liu2017AHR} & 2017 & 1,055 & 1 & 1,055 & 1105×791 & \begin{tabular}[c]{@{}c@{}} HRSC dataset includes high-resolution satellite images along with corresponding ship positions and class labels.\end{tabular} \\
 & LEVIR~\cite{Chen2020ASA} & 2020 & 37.91 & 3 & 11,028 & 800×600 & \begin{tabular}[c]{@{}c@{}} LEVIR dataset covers most types of ground features in human residential environments, such as urban, rural, and mountainous.\end{tabular} \\
 & RSOD~\cite{Sun2021RSODRS} & 2021 & 936 & 4 & 7,400 & 1051×900 & \begin{tabular}[c]{@{}c@{}}RSOD dataset includes objects such as airplanes, oil tanks, sports fields, and overpasses.\end{tabular} \\
 & Stanford~\cite{Robicquet2016LearningSE} & 2016 & 17,351 & 6 & 355,443 & 1424×1088 & \begin{tabular}[c]{@{}c@{}}Stanford Drone dataset contains trajectory and interaction information of 20,000 objects on campus in drones perspective.\end{tabular} \\
 & Visdrone~\cite{Zhu2018VisionMD} & 2018 & 6,471 & 11 & 77,547 & 1509×849 & \begin{tabular}[c]{@{}c@{}}Visdrone dataset consists of high-quality images and videos captured by UAV, along with rich object annotation information.\end{tabular} \\
 \midrule
\multirow{4}{*}{\texttt{SEG-4}} & iSAID~\cite{Zamir2019iSAIDAL} & 2019 & 30,821 & 15 & 987,239 & 896×896 & \begin{tabular}[c]{@{}c@{}}iSAID dataset consists of a large number of high spatial resolution images and includes fifteen important and common categories.\end{tabular} \\
 & loveDA~\cite{Wang2021LoveDAAR} & 2021 & 4,187 & 6 & 97,989 & 1024×1024 & \begin{tabular}[c]{@{}c@{}} LoveDA dataset consists of high-resolution images and 166,768 annotated semantic objects from 3 cities.\end{tabular} \\
 & Potsdam~\cite{potsdam} & 2012 & 5,421 & 4 & 92,161 & 512×512 & \begin{tabular}[c]{@{}c@{}}Potsdam dataset is a semantic segmentation urban remote sensing dataset and involves five foreground classes .\end{tabular} \\
 & Vaihingen~\cite{vaihingen} & 2012 & 742 & 4 & 16,875 & 512×512 & \begin{tabular}[c]{@{}c@{}}Vaihingen dataset is also used for semantic segmentation and involves the same category information as the Potsdam dataset\end{tabular} \\
 \bottomrule
\end{tabular}%
}
\end{table*}

\begin{figure}
    \centering
    \includegraphics[width=\linewidth]{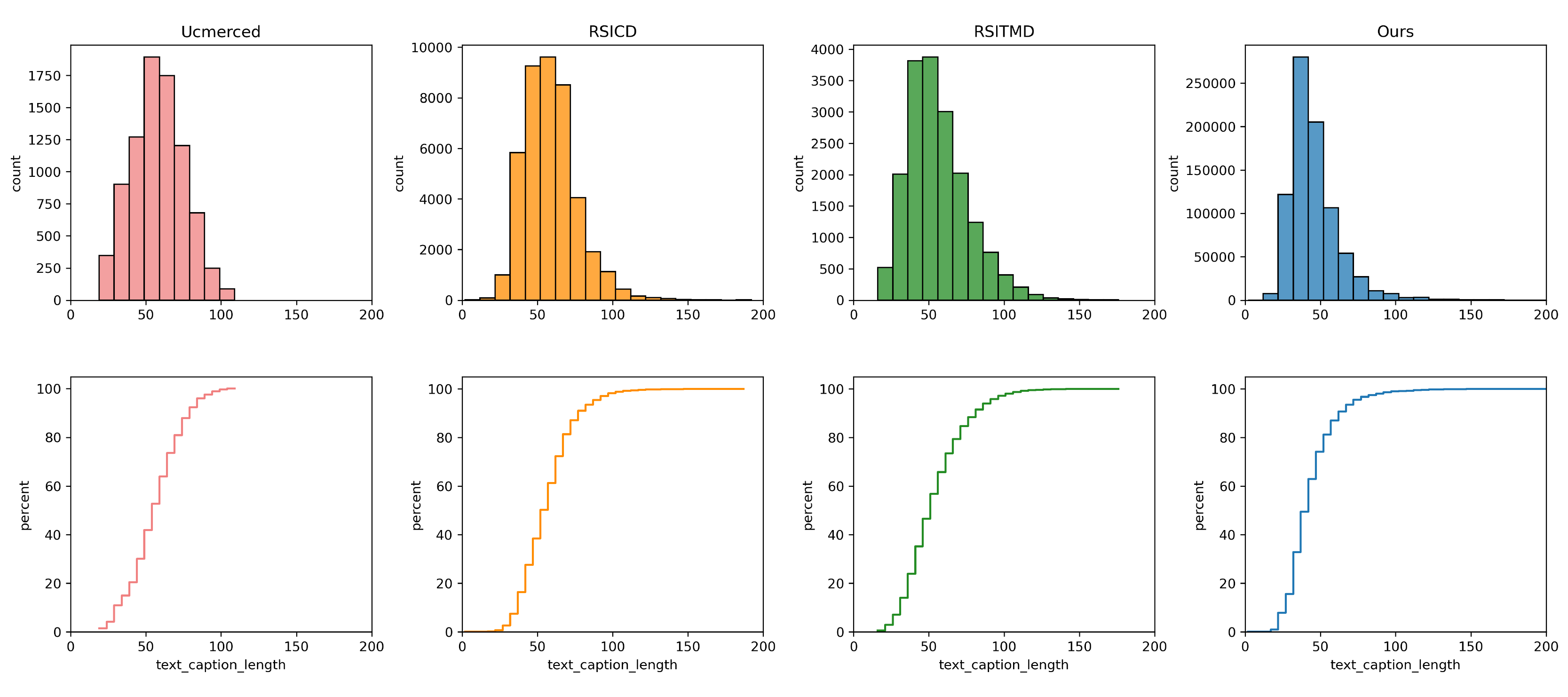}
    \caption{Distribution of caption length of existing image-text datasets UCM (pink),  RSICD (yellow), RSITMD (green), and our final dataset (blue).}
    \label{fig:caption_length}
\end{figure}

\begin{figure*}
    \centering
    \includegraphics[width=\linewidth]{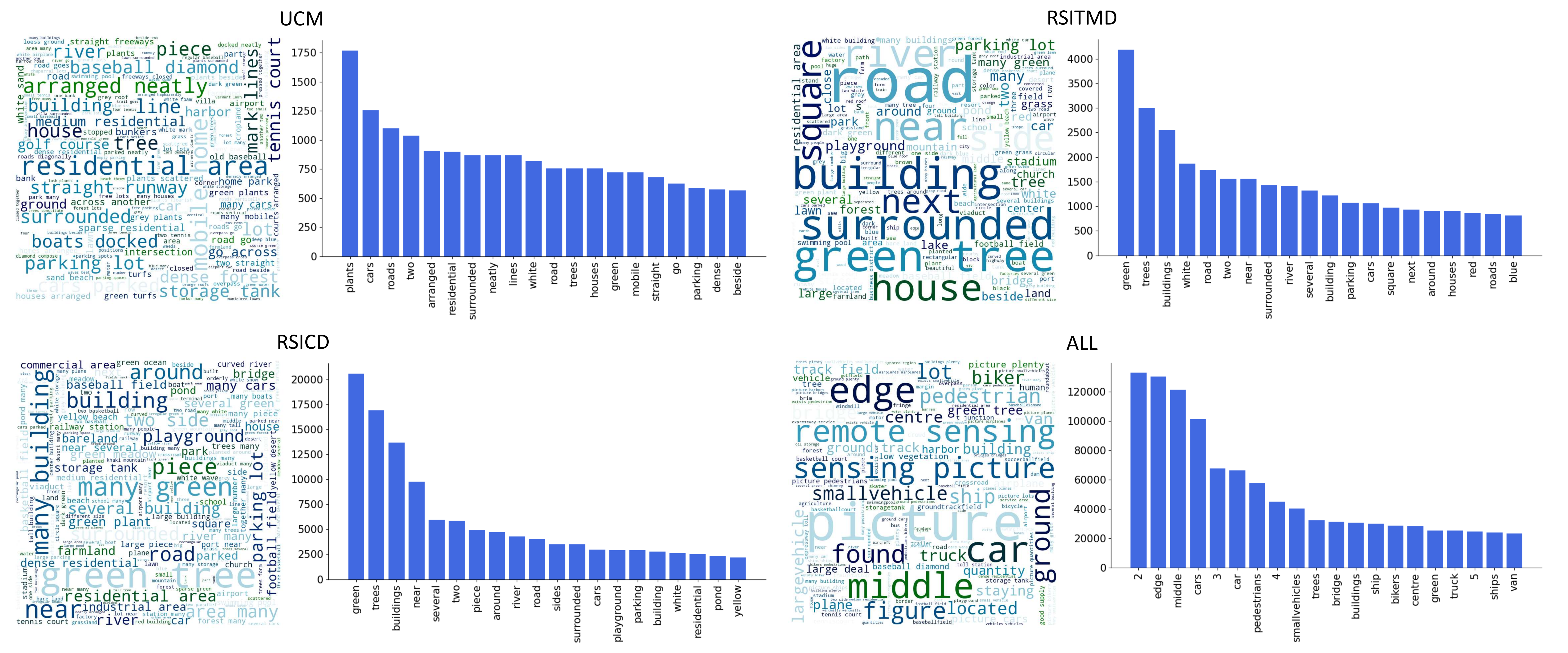}
    \caption{Word clouds and top 20 keywords of captions in existing image-text datasets UCM, RSITMD, and RSICD  and our final dataset produced by B2C and M2B from \texttt{DET-10}, \texttt{SEG-4}, and \texttt{RET-3}.}
    \label{fig:wordcloud_keyword}
\end{figure*}

\textcolor{black}{In Fig.~\ref{fig:caption_length}, we present a visualization of the caption length distribution for both the \texttt{RET-3} data and our final data. It is evident from the figure that the B2C and M2B approaches yield a caption distribution that closely mirrors that of the \texttt{RET-3} data. Furthermore, Fig.~\ref{fig:wordcloud_keyword} provides visualizations of word clouds and the top 20 keywords, with common stop-words such as ``there'', ``an'', ``is'', and others being filtered out.}

% \begin{figure}
%     \centering
%     \includegraphics[width=\linewidth]{wordcloud.pdf}
%     \caption{Word clouds of captions in existing image-text datasets UCM, RSITMD, and RSICD (upper row) and our final dataset produced by B2C and M2B from \texttt{DET-10}, \texttt{SEG-4}, and \texttt{RET-3} (bottom row).}
%     \label{fig:word_cloud}
% \end{figure}
% \begin{figure}
%     \centering
%     \includegraphics[width=\linewidth]{keyword_top_50.pdf}
%     \caption{Top 50 keywords in existing image-text datasets UCM, RSITMD, and RSICD and our final dataset.}
%     \label{fig:keyword}
% \end{figure*}

\begin{figure*}
    \centering
    \includegraphics[width=\linewidth]{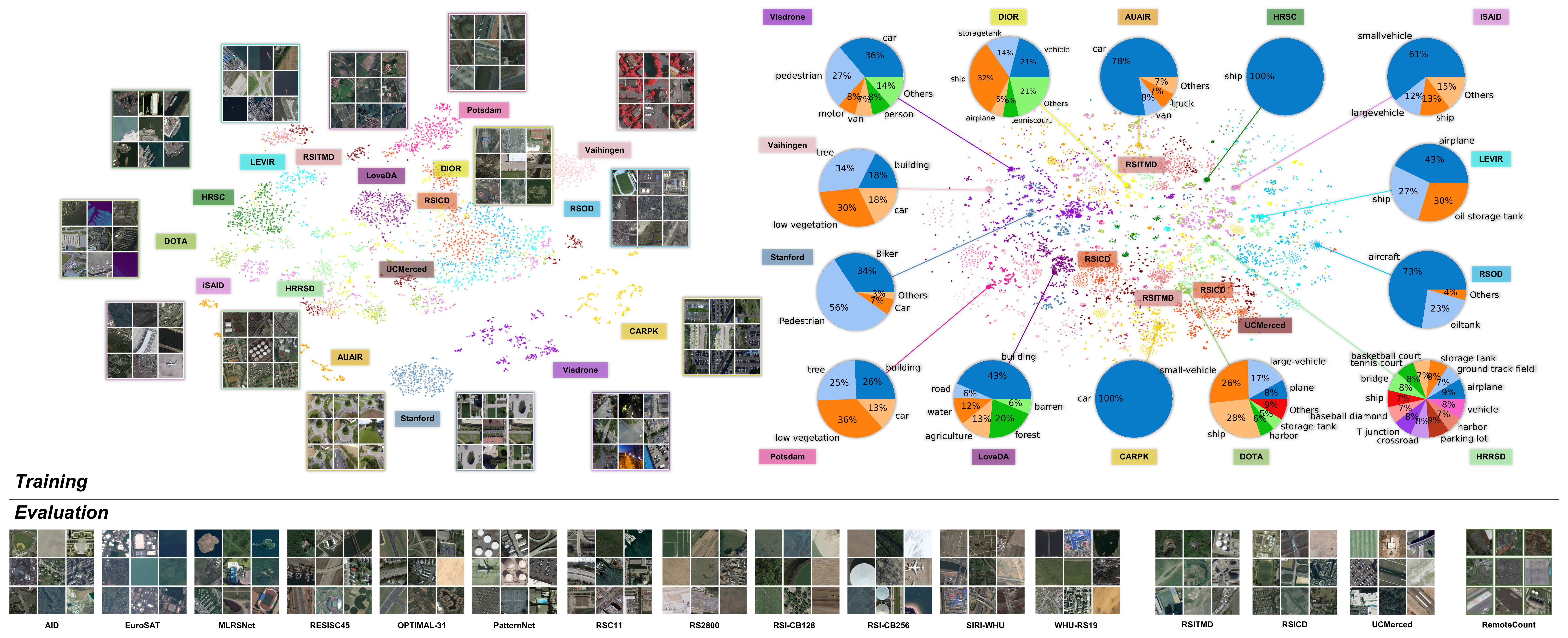}
    \caption{T-SNE visualization of image (upper left) and caption samples (upper right) in our final dataset. We provide random image samples of each dataset and label distribution of the caption samples. In the bottom row, we visualize random samples from downstream datasets used for evaluation, including 12 classification datasets, 3 retrieval datasets, and a novel object counting dataset.}
    \label{fig:tsne}
\end{figure*}

Finally, we produce a T-SNE visualization of our final data (\texttt{DET-10 + SEG-4 + Ret-3}). We select 2k samples from each subset in our final data for T-SNE visualizations. For the text T-SNE visualization, we employ paraphrase-distilroberta-base-v2 from Sentence-Transformer to extract features from textual descriptions. For the image T-SNE visualization, we simply choose ViT-Base-32 from OpenCLIP to extract visual features. From Fig.~\ref{fig:tsne}, it can be seen that our data scaling approach provides much more enriched samples. Learning multimodal representations from such a diverse sample distribution results in a strong RemoteCLIP model that handles downstream tasks in various domains.

\section{Experiments}
\label{sec:experiments}
\subsection{Implementation Details}

\subsubsection{Model} We select three types of visual backbone architecture for the RemoteCLIP model, ranging from a small-scale model ResNet-50 of 38M parameters, a medium-scale model ViT-Base-32 of 87M parameters to a large-scale model ViT-Large-14 of 304M parameters, to prove that our data scaling approach benefit different sizes of models. The ResNet-50 structure is modified from the OpenAI version. It replaces the original three 3×3 convolutions with a single 7×7 convolution and replaces the average pooling with the max pooling. Additionally, an anti-aliased rect-2 blur pooling layer is added on top of the ResNet-50 architecture, and the original average pooling layer is replaced with a multi-head self-attention-based pooling. ViT-B-32 partitions the input image into fixed-size image patches of 32×32 pixels and consists of 12 layers and 12 attention heads. ViT-L-14 partitions the input image into patches of 14×14 pixels and comprises 24 layers and 16 attention heads. The text encoder utilizes the Transformer architecture, consisting of 12 layers and 8 attention heads. The maximum token sequence length is set to 77, the same as the original OpenAI CLIP. The InfoNCE loss operates on the \texttt{[CLS]} token produced by the image and text backbone.

\subsubsection{\textcolor{black}{Data and preprocessing}} \textcolor{black}{Our final training dataset comprises a total of 165,745 images, with each image accompanied by 5 corresponding captions. This results in 828,725 training image-text pairs. For data augmentation, we utilize standard operations. For instance, we employ random crops to resize images, ensuring they align with the model's input specifications by adjusting them to the required sizes and resolutions. To enhance dataset diversity and bolster the model's robustness across various image orientations, we apply random horizontal flips, random rotations of 0$^{\circ}$, 90$^{\circ}$, 180$^{\circ}$, and 270$^{\circ}$ degrees to the images to encourage rotation invariance.}

\subsubsection{Optimization} The implementation of RemoteCLIP is based on the ITRA codebase\footnote{\url{https://itra.readthedocs.io/}} developed from OpenCLIP. We utilize automatic mixed-precision (AMP) to maintain model accuracy while reducing memory usage. Similar to CLIP, The training process is accelerated by employing the Adam optimizer.  We adopt linear warm-up and cosine learning rate scheduler. The learning rate is set to 7e-5, 4e-5, and 1e-4 respectively for ResNet-50, ViT-Base-32, and ViT-Large-14 models, and the corresponding batch size is set to 256, 256, and 28, respectively. We train all models for a total step of 108,215. Using a single-node 4$\times$NVIDIA 3090Ti machine, training our largest RemoteCLIP model takes 233.4 hours.

\subsection{Benchmarking RemoteCLIP}

\begin{table*}[t]
\centering
\caption{Cross-modal retrieval performance on RSITMD, RSICD and UCM benchmarks. CLIP-CL corresponds to the continual pretraining of CLIP on existing \texttt{RET-3} data only (details presented in Section~\ref{sec:clip_cl}). Previous SOTA results are marked by \textcolor{red}{RED}, while the RemoteCLIP models are marked by \textcolor{blue}{BLUE}. Our RemoteCLIP achieves the SOTA performance on all three retrieval benchmarks by a large margin compared to previous approaches.}
\label{tab:retrieval}
\resizebox{\textwidth}{!}{%
\begin{tabular}{ccccccccccccccccc}

\toprule
\multirow{2}{*}{\textbf{{\makecell[c]{Testing \\ Dataset}}}} & \multirow{2}{*}{\textbf{{\makecell[c]{Training \\ Dataset}}}} & \multirow{2}{*}{\textbf{\makecell[c]{Training\\ Samples}}} & \multirow{2}{*}{\textbf{Date}}& \multirow{2}{*}{\textbf{Method}} & \multicolumn{2}{c}{\textbf{Image Backbone}} & \multicolumn{2}{c}{\textbf{Text Backbone}} & \multirow{2}{*}{\textbf{{\makecell[c]{Total \\ Params}}}} & \multicolumn{3}{c}{\textbf{Image to Text}} & \multicolumn{3}{c}{\textbf{Text to Image}} & \multirow{2}{*}{\textbf{{\makecell[c]{Mean \\ Recall}}}} \\
\cline{6-16}
 &  &  &  &  & \textbf{Name} & \textbf{Params} & \textbf{Name} & \textbf{Params} &  & \textbf{R@1} & \textbf{R@5} & \textbf{R@10} & \textbf{R@1} & \textbf{R@5} & \textbf{R@10} &  \\
\midrule
\multirow{18}{*}{RSITMD} & \multirow{13}{*}{RSITMD} & \multirow{13}{*}{4,743} & Jul 2017 & VSE++~\cite{Faghri2017VSEIV} & VGG19 & - & GRU & - & - & 10.38 & 27.65 & 39.60 & 7.79 & 24.87 & 38.67 & 24.83 \\
 &  &  & Mar 2018 & SCAN~\cite{Lee2018StackedCA} & ResNet-101 & - & GRU & - & - & 11.06 & 25.88 & 39.38 & 9.82 & 29.38 & 42.12 & 26.28 \\
 &  &  & Aug 2019 & MTFN~\cite{Wang2019MatchingIA} & ResNet & - & GRU & - & - & 10.40 & 27.65 & 36.28 & 9.96 & 31.37 & 45.84 & 26.92 \\
 &  &  & Aug 2020 & AMFMN~\cite{Hoxha2020TowardRS} & ResNet-50 & - & GloVe fasText & - & -& 10.63 & 24.78 & 41.81 & 11.51 & 34.69 & 54.87 & 29.72 \\
 &  &  & Dec 2020  & LW-MRC-u~\cite{AlRahhal2020DeepUE} & Big Transfer & - & Bi-LSTM & - & - & 9.73 & 26.77 & 37.61 & 9.25 & 34.07 & 54.03 & 28.58 \\
 &  &  & Apr 2022 & GaLR~\cite{Yuan2022RemoteSC} & ResNet-18 & - & GRU & - & - & 14.82 & 31.64 & 42.48 & 11.15 & 36.68 & 51.68 & 31.41 \\
 &  &  &  Dec 2022 & CMFM-Net~\cite{Yu2023Text} & ResNet-18 & - & GRU & - & - & 10.84 & 28.76 & 40.04 & 10.00 & 32.83 & 47.21 & 28.28 \\
 &  &  & Dec 2022 & HyperMatch~\cite{Yao2023Hypergraph} & ResNet-18 & - & GRU & - & - & 11.73 & 28.10 & 38.05 & 9.16 & 32.31 & 46.64 & 27.66 \\
 &  &  & \cellcolor[HTML]{fcddda} Oct 2022 &\cellcolor[HTML]{fcddda} Rahhal et al.~\cite{AlRahhal2022MultilanguageTF} &\cellcolor[HTML]{fcddda} ViT-B-32 &\cellcolor[HTML]{fcddda} 87 &\cellcolor[HTML]{fcddda} Transformer &\cellcolor[HTML]{fcddda} 63 &\cellcolor[HTML]{fcddda} 151 &\cellcolor[HTML]{fcddda} 19.69 &\cellcolor[HTML]{fcddda} 40.26 &\cellcolor[HTML]{fcddda} 54.42 &\cellcolor[HTML]{fcddda} 17.61 &\cellcolor[HTML]{fcddda} 49.73 &\cellcolor[HTML]{fcddda} 66.59 &\cellcolor[HTML]{fcddda} 41.38 \\
&  &  & Sept 2023 & HVSA~\cite{Zhang2023Hypersphere} & ResNet-18 & - & GRU & - & - & 13.20 & 32.08 & 45.58 & 11.43 & 39.20 & 57.45 & 33.16 \\
&  &  & May 2022 & FBCLM~\cite{Li2022fusion} & Resnet-18 & - & BERT & - & - & 12.84 & 30.53 & 45.89 & 10.44 & 37.01 & 57.94 & 32.44 \\
&  &  & Oct 2023 & DOVE~\cite{Ma2023Direction} & ResNet-50 & - & GRU & - & - & 16.81 & 36.80 & 49.93 & 12.20 & 49.93 & 66.50 & 38.70 \\
&  &  & Oct 2023 & PIR~\cite{Pan2023Prior} & SwinT+ ResNet-50 & - & Bert & - & - & 18.14 & 41.15 & 52.88 & 12.17 & 41.68 & 63.41 & 38.24 \\
 \cline{2-17}
 & \multirow{2}{*}{\texttt{RET-3}} & \multirow{2}{*}{13,713}  & - & CLIP-CL & ResNet-50 & 38 & Transformer & 64 & 102 & 19.25 & 39.82 & 51.33 & 15.09 & 41.46 & 56.64 & 37.27 \\
 &  & & - &  CLIP-CL & ViT-B-32 & 88 & Transformer & 64 & 151 & 24.78 & 50.00 & 63.27 & 22.61 & 55.27 & 69.87 & 47.63 \\
 \cline{2-17}
 & \multirow{3}{*}{\texttt{RET-3} + \texttt{DET-10} + \texttt{SEG-4}} & \multirow{3}{*}{165,745} &  \cellcolor[HTML]{DAE8FC} Jun 2023 &\cellcolor[HTML]{DAE8FC} RemoteCLIP &\cellcolor[HTML]{DAE8FC} ResNet-50 &\cellcolor[HTML]{DAE8FC} 38 &\cellcolor[HTML]{DAE8FC} Transformer &\cellcolor[HTML]{DAE8FC} 64 &\cellcolor[HTML]{DAE8FC} 102 &\cellcolor[HTML]{DAE8FC} 23.67 &\cellcolor[HTML]{DAE8FC} 47.57 &\cellcolor[HTML]{DAE8FC} 64.60 &\cellcolor[HTML]{DAE8FC} 19.29 &\cellcolor[HTML]{DAE8FC} 51.55 &\cellcolor[HTML]{DAE8FC} 70.58 &\cellcolor[HTML]{DAE8FC} 46.21 \\
 &  &  &\cellcolor[HTML]{DAE8FC}Jun 2023 &\cellcolor[HTML]{DAE8FC} RemoteCLIP &\cellcolor[HTML]{DAE8FC} ViT-B-32 &\cellcolor[HTML]{DAE8FC} 87 &\cellcolor[HTML]{DAE8FC} Transformer &\cellcolor[HTML]{DAE8FC} 63 &\cellcolor[HTML]{DAE8FC} 151 &\cellcolor[HTML]{DAE8FC} 27.88 &\cellcolor[HTML]{DAE8FC} 50.66 &\cellcolor[HTML]{DAE8FC} \textbf{65.71} &\cellcolor[HTML]{DAE8FC} 22.17 &\cellcolor[HTML]{DAE8FC} 56.46 &\cellcolor[HTML]{DAE8FC} 73.41 &\cellcolor[HTML]{DAE8FC} 49.38 \\
 &  &  &\cellcolor[HTML]{DAE8FC} Jun 2023 &\cellcolor[HTML]{DAE8FC} RemoteCLIP &\cellcolor[HTML]{DAE8FC} ViT-L-14 &\cellcolor[HTML]{DAE8FC} 304 &\cellcolor[HTML]{DAE8FC} Transformer &\cellcolor[HTML]{DAE8FC} 124 &\cellcolor[HTML]{DAE8FC} 428 &\cellcolor[HTML]{DAE8FC} \textbf{28.76} &\cellcolor[HTML]{DAE8FC} \textbf{52.43} &\cellcolor[HTML]{DAE8FC} 63.94 &\cellcolor[HTML]{DAE8FC} \textbf{23.76} &\cellcolor[HTML]{DAE8FC} \textbf{59.51} &\cellcolor[HTML]{DAE8FC} \textbf{74.73} &\cellcolor[HTML]{DAE8FC} \textbf{50.52} \\
 \midrule
\multirow{19}{*}{RSICD} & \multirow{14}{*}{RSICD} & \multirow{14}{*}{10,921} & Jul 2017 & VSE++~\cite{Faghri2017VSEIV} & VGG19 & - & GRU & - & - & 3.38 & 9.51 & 17.46 & 2.82 & 11.32 & 18.10 & 10.43 \\
 &  &  & Mar 2018 & SCAN~\cite{Lee2018StackedCA} & ResNet-101 & - & GRU & - & - & 5.85 & 12.89 & 19.84 & 3.71 & 16.40 & 26.73 & 14.23 \\
 &  &  & Aug 2019 & MTFN~\cite{Wang2019MatchingIA} & ResNet & - & GRU & - & - & 5.02 & 12.52 & 19.74 & 4.90 & 17.17 & 29.49 & 14.81 \\
 &  &  & Aug 2020 & AMFMN~\cite{Hoxha2020TowardRS} & ResNet-50 & - & GloVe fasText & - & - & 5.39 & 15.08 & 23.40 & 4.90 & 18.28 & 31.44 & 16.42 \\
 &  &  & Dec 2020 & LW-MRC-u~\cite{AlRahhal2020DeepUE} & Big Transfer & - & Bi-LSTM & - & - & 4.39 & 13.35 & 20.29 & 4.30 & 18.85 & 32.34 & 15.59 \\
 &  &  & Apr 2022 & GaLR~\cite{Yuan2022RemoteSC} & ResNet-18 & - & GRU & - & - & 6.59 & 19.85 & 31.04 & 4.69 & 19.48 & 32.13 & 18.96 \\
 &  &  & Dec 2022 & CMFM-Net~\cite{Yu2023Text} &ResNet-18 & - & GRU & - & - & 5.40 & 18.66 & 28.55 & 5.31 & 18.57 & 30.03 & 17.75 \\
&  &  & Dec 2022 & HyperMatch~\cite{Yao2023Hypergraph} &  ResNet-18 & - & GRU & - & - & 7.14 & 20.04 & 31.02 & 6.08 & 20.37 & 33.82 & 19.75 \\
&  &  & Sept 2022 & KCR~\cite{Mi2022Knowledge} & ResNet-101 & - & Transformer & - & -  & 5.84 & 22.31 & 36.12 & 4.76 & 18.59 & 27.20 & 19.14 \\
&  &  & \cellcolor[HTML]{fcddda} Oct 2022 &\cellcolor[HTML]{fcddda} Rahhal et al.~\cite{AlRahhal2022MultilanguageTF} &\cellcolor[HTML]{fcddda} ViT-B-32 &\cellcolor[HTML]{fcddda} 87 &\cellcolor[HTML]{fcddda} Transformer &\cellcolor[HTML]{fcddda} 63 &\cellcolor[HTML]{fcddda} 151 &\cellcolor[HTML]{fcddda} 10.70 &\cellcolor[HTML]{fcddda} 29.64 &\cellcolor[HTML]{fcddda} 41.53 &\cellcolor[HTML]{fcddda} 9.14 &\cellcolor[HTML]{fcddda} 28.96 &\cellcolor[HTML]{fcddda} 44.59 &\cellcolor[HTML]{fcddda} 27.43 \\
&  &  & Sept 2023 & HVSA~\cite{Zhang2023Hypersphere} &  ResNet-18 & - & GRU & - & - & 7.47 & 20.62 & 32.11 & 5.51 & 21,13 & 34,13 & 16.43 \\
&  &  & May 2022 & FBCLM~\cite{Li2022fusion} & Resnet-18 & - & BERT & - & - & 13.27 & 27.17 & 37.60 & 13.54 & 38.74 & 56.94 & 31.21 \\
&  &  & Oct 2023 & DOVE~\cite{Ma2023Direction} & ResNet-50 & - & GRU & - & -  & 8.66 & 22.35 & 34.95 & 6.04 & 23.95 & 40.35 & 22.72 \\
&  &  & Oct 2023 & PIR~\cite{Pan2023Prior} & SwinT + ResNet-50 &  & Bert &  &  & 9.88 & 27.26 & 39.16 & 6.97 & 24.56 & 38.92 & 24.46 \\
 \cline{2-17}
 & \multirow{2}{*}{\texttt{RET-3}} & \multirow{2}{*}{13,713} &- & CLIP-CL & ResNet-50 & 38 & Transformer & 64 & 102 & 12.99 & 26.35 & 36.32 & 8.56 & 25.60 & 39.16 & 24.83 \\
 &  & &- & CLIP-CL & ViT-B-32 & 88 & Transformer & 64 & 151 & 17.84 & 35.96 & 50.14 & 13.89 & 35.15 & 50.08 & 33.96 \\
 \cline{2-17}
 & \multirow{3}{*}{\texttt{RET-3} + \texttt{DET-10} + \texttt{SEG-4}} & \multirow{3}{*}{165,745} & \cellcolor[HTML]{DAE8FC} Jun 2023 &\cellcolor[HTML]{DAE8FC} RemoteCLIP &\cellcolor[HTML]{DAE8FC} ResNet-50 &\cellcolor[HTML]{DAE8FC} 38 &\cellcolor[HTML]{DAE8FC} Transformer &\cellcolor[HTML]{DAE8FC} 64 &\cellcolor[HTML]{DAE8FC} 102 &\cellcolor[HTML]{DAE8FC} 13.36 &\cellcolor[HTML]{DAE8FC} 32.94 &\cellcolor[HTML]{DAE8FC} 44.83 &\cellcolor[HTML]{DAE8FC} 10.76 &\cellcolor[HTML]{DAE8FC} 32.83 &\cellcolor[HTML]{DAE8FC} 48.75 &\cellcolor[HTML]{DAE8FC} 30.58 \\
 &  &  &\cellcolor[HTML]{DAE8FC} Jun 2023 &\cellcolor[HTML]{DAE8FC} RemoteCLIP &\cellcolor[HTML]{DAE8FC} ViT-B-32 &\cellcolor[HTML]{DAE8FC} 87 &\cellcolor[HTML]{DAE8FC} Transformer &\cellcolor[HTML]{DAE8FC} 63 &\cellcolor[HTML]{DAE8FC} 151 &\cellcolor[HTML]{DAE8FC} 17.02 &\cellcolor[HTML]{DAE8FC} \textbf{37.97} &\cellcolor[HTML]{DAE8FC} \textbf{51.51} &\cellcolor[HTML]{DAE8FC} 13.71 &\cellcolor[HTML]{DAE8FC} 37.11 &\cellcolor[HTML]{DAE8FC} 54.25 &\cellcolor[HTML]{DAE8FC} 35.26 \\
 &  &  &\cellcolor[HTML]{DAE8FC} Jun 2023 &\cellcolor[HTML]{DAE8FC} RemoteCLIP &\cellcolor[HTML]{DAE8FC} ViT-L-14 &\cellcolor[HTML]{DAE8FC} 304 &\cellcolor[HTML]{DAE8FC} Transformer &\cellcolor[HTML]{DAE8FC} 124 &\cellcolor[HTML]{DAE8FC} 428 &\cellcolor[HTML]{DAE8FC} \textbf{18.39} &\cellcolor[HTML]{DAE8FC} 37.42 &\cellcolor[HTML]{DAE8FC} 51.05 &\cellcolor[HTML]{DAE8FC} \textbf{14.73} &\cellcolor[HTML]{DAE8FC} \textbf{39.93} &\cellcolor[HTML]{DAE8FC} \textbf{56.58} &\cellcolor[HTML]{DAE8FC} \textbf{36.35} \\
 \midrule
\multirow{12}{*}{UCM} & \multirow{7}{*}{UCM} & \multirow{7}{*}{2,100} & Jul 2017 & VSE++~\cite{Faghri2017VSEIV} & VGG19 & - & GRU & - & - & 12.38 & 44.76 & 65.71 & 10.10 & 31.80 & 56.85 & 36.93 \\
 &  &  & Mar 2018 & SCAN~\cite{Lee2018StackedCA} & ResNet-101 & - & GRU & - & - & 12.85 & 47.14 & 69.52 & 12.48 & 46.86 & 71.71 & 43.43 \\
 &  &  & Aug 2019 & MTFN~\cite{Wang2019MatchingIA} & ResNet & - & GRU & - & - & 10.47 & 47.62 & 64.29 & 14.19 & 52.38 & 78.95 & 44.65 \\
 &  &  & Aug 2020 & AMFMN~\cite{Hoxha2020TowardRS} & ResNet-50 & - & GloVe fasText & - & - & 16.67 & 45.71 & 68.57 & 12.86 & 53.24 & 79.43 & 46.08 \\
 &  &  & Dec 2020 & LW-MRC-u~\cite{AlRahhal2020DeepUE} & Big Transfer & - & Bi-LSTM & - & - & 18.10 & 47.14 & 63.81 & 13.14 & 50.38 & 79.52 & 45.35 \\
 &  &  & \cellcolor[HTML]{fcddda} Oct 2022 &\cellcolor[HTML]{fcddda} Rahhal et al.~\cite{AlRahhal2022MultilanguageTF} &\cellcolor[HTML]{fcddda} ViT-B-32 &\cellcolor[HTML]{fcddda} 87 &\cellcolor[HTML]{fcddda} Transformer &\cellcolor[HTML]{fcddda} 63 &\cellcolor[HTML]{fcddda} 151 &\cellcolor[HTML]{fcddda} 19.04 &\cellcolor[HTML]{fcddda} 53.33 &\cellcolor[HTML]{fcddda} 77.61 &\cellcolor[HTML]{fcddda} \textbf{19.33} &\cellcolor[HTML]{fcddda} \textbf{64.00} &\cellcolor[HTML]{fcddda} 91.42 &\cellcolor[HTML]{fcddda} 54.12 \\
 &  &  & May 2022 & FBCLM~\cite{Li2022fusion} &Resnet-18 & - & BERT & - & -  & 28.57 & 63.81 & 82.86 & 27.33 & 72.67 & 94.38 & 61.60 \\
 \cline{2-17}
 & \multirow{2}{*}{\texttt{RET-3}} & \multirow{2}{*}{13,713} & - & CLIP-CL & ResNet-50 & 38 & Transformer & 64 & 102 & 14.29 & 44.76 & 72.86 & 13.52 & 53.71 & 82.10 & 46.87 \\
& & & - & CLIP-CL & ViT-B-32 & 88 & Transformer & 64 & 151 & 20.00 & 55.24 & 80.95 & 18.19 & 64.86 & 92.38 & 55.27 \\
 \cline{2-17}
 & \multirow{3}{*}{\texttt{RET-3} + \texttt{DET-10} + \texttt{SEG-4}} & \multirow{3}{*}{165,745} & \cellcolor[HTML]{DAE8FC} 2023  &\cellcolor[HTML]{DAE8FC} RemoteCLIP &\cellcolor[HTML]{DAE8FC} ResNet-50 &\cellcolor[HTML]{DAE8FC} 38 &\cellcolor[HTML]{DAE8FC} Transformer &\cellcolor[HTML]{DAE8FC} 64 &\cellcolor[HTML]{DAE8FC} 102 &\cellcolor[HTML]{DAE8FC} 13.33 &\cellcolor[HTML]{DAE8FC} 50.48 &\cellcolor[HTML]{DAE8FC} 74.76 &\cellcolor[HTML]{DAE8FC} 15.24 &\cellcolor[HTML]{DAE8FC} 57.14 &\cellcolor[HTML]{DAE8FC} 84.57 &\cellcolor[HTML]{DAE8FC} 49.25 \\
 &  &  &\cellcolor[HTML]{DAE8FC} 2023  &\cellcolor[HTML]{DAE8FC} RemoteCLIP &\cellcolor[HTML]{DAE8FC} ViT-B-32 &\cellcolor[HTML]{DAE8FC} 87 &\cellcolor[HTML]{DAE8FC} Transformer &\cellcolor[HTML]{DAE8FC} 63 &\cellcolor[HTML]{DAE8FC} 151 &\cellcolor[HTML]{DAE8FC} \textbf{20.48} &\cellcolor[HTML]{DAE8FC} \textbf{59.85} &\cellcolor[HTML]{DAE8FC} \textbf{83.33} &\cellcolor[HTML]{DAE8FC} 18.67 &\cellcolor[HTML]{DAE8FC} 61.52 &\cellcolor[HTML]{DAE8FC} \textbf{94.29} &\cellcolor[HTML]{DAE8FC} \textbf{56.36} \\
 &  &  &\cellcolor[HTML]{DAE8FC} Jun 2023 &\cellcolor[HTML]{DAE8FC} RemoteCLIP &\cellcolor[HTML]{DAE8FC} ViT-L-14 &\cellcolor[HTML]{DAE8FC} 304 &\cellcolor[HTML]{DAE8FC} Transformer &\cellcolor[HTML]{DAE8FC} 124 &\cellcolor[HTML]{DAE8FC} 428 &\cellcolor[HTML]{DAE8FC} 19.05 &\cellcolor[HTML]{DAE8FC} 54.29 &\cellcolor[HTML]{DAE8FC} 80.95 &\cellcolor[HTML]{DAE8FC} 17.71 &\cellcolor[HTML]{DAE8FC} 62.19 &\cellcolor[HTML]{DAE8FC} 93.90 &\cellcolor[HTML]{DAE8FC} 54.68 \\
\bottomrule
\end{tabular}%
}
\end{table*}

\subsubsection{Cross-modal Retrieval}

We first present the performance of RemoteCLIP on three remote sensing image-text retrieval benchmarks (RSITMD, RSICD, UCM) and compare it with previous results. To perform cross-modal retrieval with RemoteCLIP, we extract image and text representations on the test split, perform L-2 normalization, and retrieve most similar samples based on the dot-product similarity measure. We report the retrieval recall of top-1 (R@1), top-5 (R@5), top-10 (R@10), and the mean recall of these values. We do not perform any dataset-specific fine-tuning or re-ranking to improve the results.

Table~\ref{tab:retrieval} summarizes the results. We also provide the details of each model, including training data, backbone architecture, and the number of parameters (in million), for better comparison. Our RemoteCILP model achieves SOTA performance on all three retrieval benchmarks. On the challenging RSITMD and RSICD datasets, our model outperforms the previous SOTA method (Rahhal et al.~\cite{AlRahhal2022MultilanguageTF}) by a large margin (9.14\% and 8.92\% respectively). Such results are achieved with the large-scale model ViT-Large-14 with 304 parameters. When it comes to smaller models, RemoteCLIP is still competitive -- the ResNet-50-based RemoteCLIP can also exceed previous SOTA methods on RSITMD and RSICD datasets. In addition, on all three retrieval benchmarks, RemoteCLIP outperforms the \texttt{CLIP-CL} baseline which only uses the existing \texttt{RET-3} data, showing the effectiveness of RemoteCLIP data scaling.

\subsubsection{Object Counting}

\begin{figure*}
    \centering
    \includegraphics[width=\linewidth]{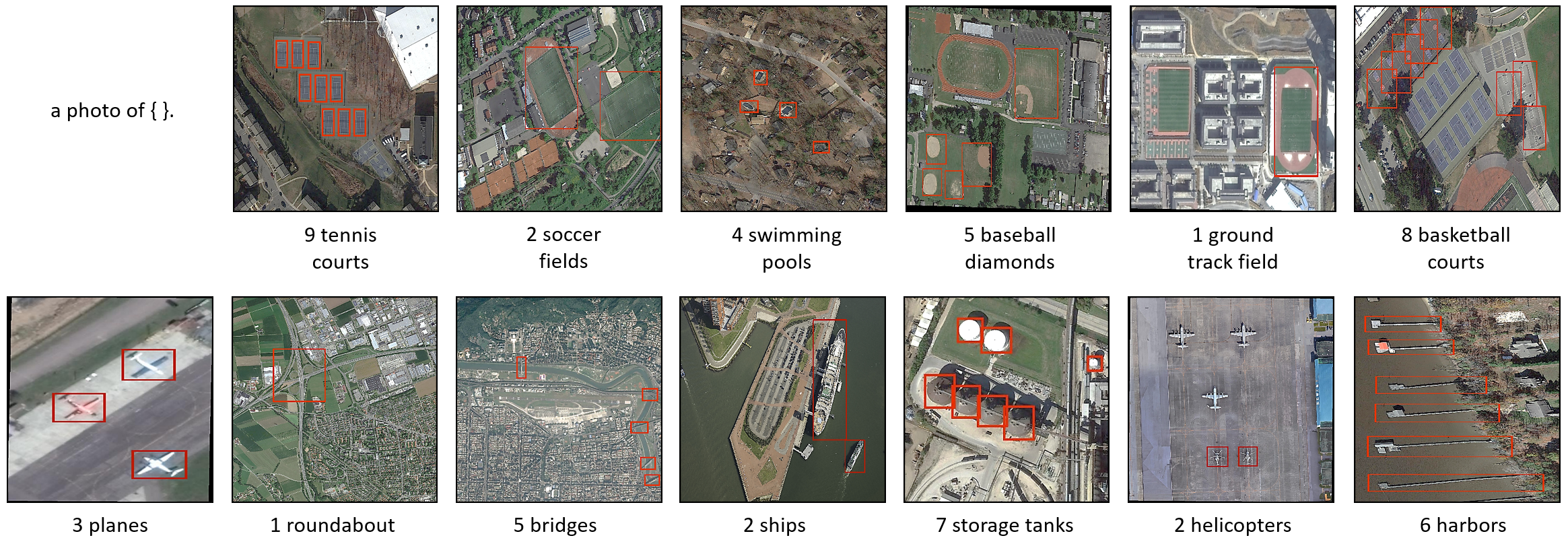}
    \caption{Visualization of the RemoteCount dataset samples. Objects of interest are annotated by red bounding boxes.}
    \label{fig:remotecount_samples}
\end{figure*}

A recent study shows that large-scale pretraining empowers the CLIP model to do zero-shot object counting~\cite{Paiss2023Teaching}. Here we are interested in whether RemoteCLIP has such fine-grained language understanding capability. To answer this question, we introduce a new remote sensing counting benchmark ``RemoteCount" to evaluate the accuracy of object counting from 1 to 10. This dataset consists of 947 image-text pairs, which are mainly selected from the validation set of the DOTA dataset. It covers 13 categories, including planes, helicopters, roundabouts, bridges, baseball diamonds, ground track fields, basketball courts, tennis courts, harbors, soccer fields, swimming pools,  ships, and storage tanks. The dataset is annotated by five graduate students, careful manual verification is conducted to ensure its quality. Fig.~\ref{fig:remotecount_samples} visualize random samples within RemoteCount.

We focus on comparing the zero-shot counting accuracy of CLIP and RemoteCLIP. For each image, we augment the existing caption with nine other possible captions by replacing the number in its caption with all the numbers from 1 to 10 and calculate the similarity score between the image and each of the ten captions. The number in the caption that obtains the highest similarity score with the image is considered the predicted number. 

\begin{figure}
    \centering
    \includegraphics[width=\linewidth]{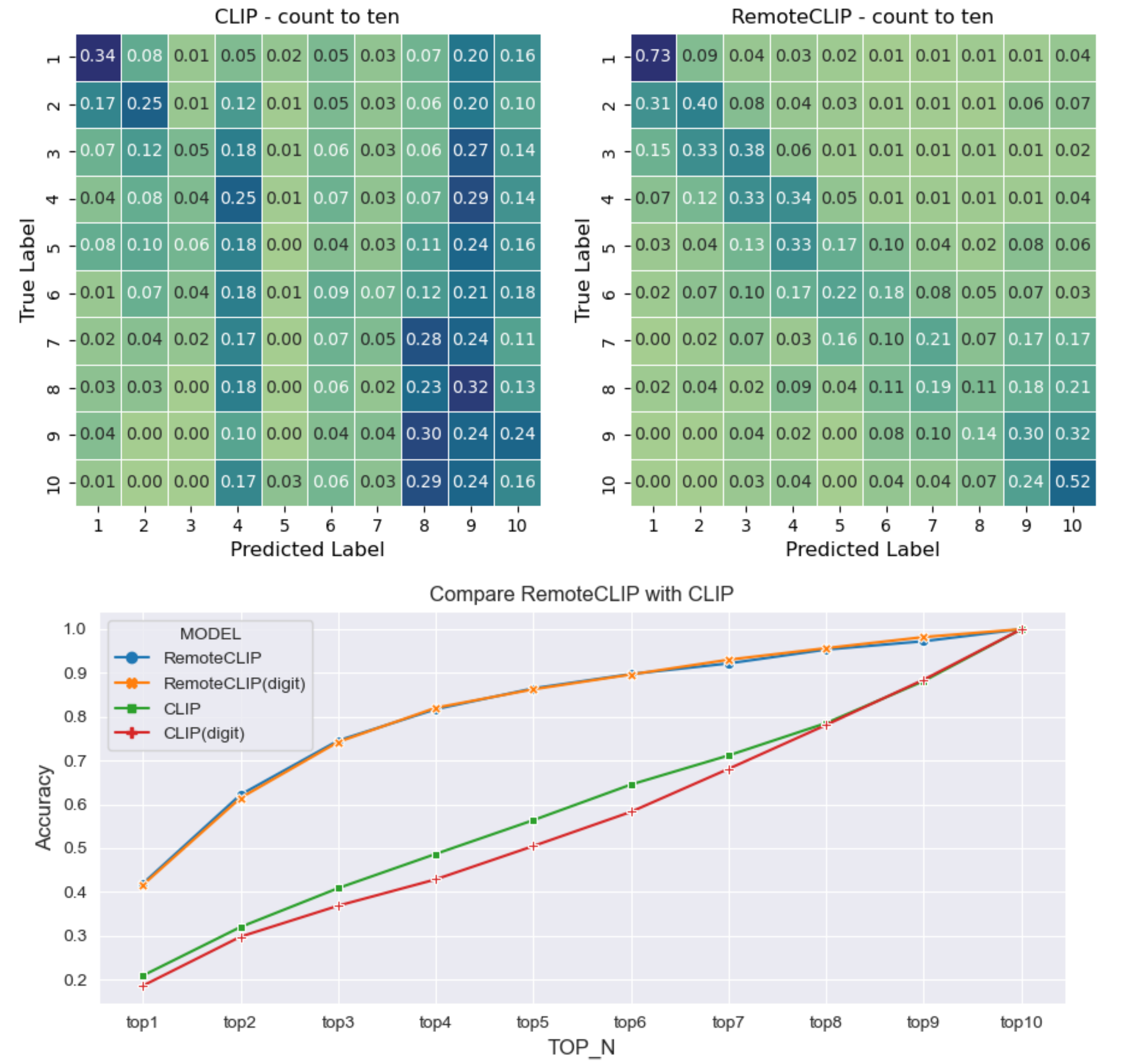}
    \caption{The object counting experiment of CLIP and RemoteCLIP on RemoteCount dataset. Upper row: The confusion matrix of CLIP and RemoteCLIP.  Bottom row: Top-1 accuracy to top-10 accuracy of CLIP and RemoteCLIP. }
    \label{fig:counting_results}
\end{figure}

The evaluation results are provided in Fig.~\ref{fig:counting_results}. The top row shows the confusion matrix of CLIP and RemoteCLIP. We normalize the final output confusion matrix. CLIP has shown poor results in this task, but RemoteCLIP has a clear diagonal representing its higher accuracy. We also present the top 1 to top 10 accuracy of CLIP and RemoteCLIP. It is clear that at the top 6 accuracy level, RemoteCLIP still significantly outperforms CLIP. In addition, we also test a variation of replacing the number with ``1''-``10'' instead of ``one''-``ten'', which we denote as ``(digit)'' in Fig.~\ref{fig:counting_results}. As can be seen, RemoteCLIP is much more robust to such variation.

\subsubsection{Zero-shot Image Classification}
\label{section:zero-shot}

\begin{table*}[]
\centering
\caption{Zero-shot accuracies on 12 remote sensing image classification datasets.}
\label{tab: zero-shot}
\resizebox{0.85\linewidth}{!}{%
\begin{tabular}{ccccccccccccccc}
% \toprule
\textbf{Method} & \textbf{Backbone} & \rotatebox{90}{\begin{tabular}[c]{@{}l@{}}\textbf{RSI-CB128}\end{tabular}} & \rotatebox{90}{\begin{tabular}[c]{@{}l@{}}\textbf{RSI-CB256}\end{tabular}} & \rotatebox{90}{\begin{tabular}[c]{@{}l@{}}\textbf{WHU-earth}\end{tabular}} & \rotatebox{90}{\begin{tabular}[c]{@{}l@{}}\textbf{EuroSAT}\end{tabular}} & \rotatebox{90}{\begin{tabular}[c]{@{}l@{}}\textbf{MLRSNet}\end{tabular}} & \rotatebox{90}{\begin{tabular}[c]{@{}l@{}}\textbf{PatternNet}\end{tabular}} & \rotatebox{90}{\begin{tabular}[c]{@{}l@{}}\textbf{RESISC45}\end{tabular}} & \rotatebox{90}{\begin{tabular}[c]{@{}l@{}}\textbf{AID}\end{tabular}} & \rotatebox{90}{\begin{tabular}[c]{@{}l@{}}\textbf{RS2800}\end{tabular}} & \rotatebox{90}{\begin{tabular}[c]{@{}l@{}}\textbf{OPTIMAL-31}\end{tabular}} & \rotatebox{90}{\begin{tabular}[c]{@{}l@{}}\textbf{RSC11}\end{tabular}} & \rotatebox{90}{\begin{tabular}[c]{@{}l@{}}\textbf{WHU-RS19}\end{tabular}} & \rotatebox{90}{\begin{tabular}[c]{@{}l@{}}\textbf{Average}\end{tabular}} \\
\midrule
CLIP & \multirow{2}{*}{ResNet-50} & \textbf{26.56} & \textbf{34.24} & 40.42 & \textbf{42.02} & \textbf{45.54} & \textbf{46.96} & \textbf{53.57} & 57.35 & 62.14 & 64.52 & 64.54 & 69.42 & 50.61 \\
RemoteCLIP &  & \cellcolor[HTML]{DAE8FC} 13.95 &\cellcolor[HTML]{DAE8FC} 33.03 &\cellcolor[HTML]{DAE8FC} \textbf{56.25} &\cellcolor[HTML]{DAE8FC} 17.19 &\cellcolor[HTML]{DAE8FC} 40.68 &\cellcolor[HTML]{DAE8FC} 45.51 &\cellcolor[HTML]{DAE8FC} 53.24 &\cellcolor[HTML]{DAE8FC} \textbf{86.55} &\cellcolor[HTML]{DAE8FC} \textbf{62.86} &\cellcolor[HTML]{DAE8FC} \textbf{70.16} &\cellcolor[HTML]{DAE8FC} \textbf{66.93} &\cellcolor[HTML]{DAE8FC} \textbf{95.15} &\cellcolor[HTML]{DAE8FC} \textbf{53.46 } \\
 &  $\pm\Delta$ & {\color[HTML]{FF0000} -12.61} & {\color[HTML]{FF0000} -1.21} & {\color[HTML]{00B050} +15.83} & {\color[HTML]{FF0000} -24.83} & {\color[HTML]{FF0000} -4.86} & {\color[HTML]{FF0000} -1.45} & {\color[HTML]{FF0000} -0.33} & {\color[HTML]{00B050} +29.20} & {\color[HTML]{00B050} +0.72} & {\color[HTML]{00B050} +5.64} & {\color[HTML]{00B050} +2.39} & {\color[HTML]{00B050} +25.73} & {\color[HTML]{00B050} +2.85} \\
\midrule
CLIP & \multirow{2}{*}{ViT-B-32} & \textbf{28.88} & 37.35 & 51.18 & \textbf{47.11} & 55.29 & \textbf{58.95} & 60.92 & 65.65 & 59.31 & 68.62 & 58.35 & 80.61 & 56.02\\
RemoteCLIP &  & \cellcolor[HTML]{DAE8FC} 24.18 &\cellcolor[HTML]{DAE8FC} \textbf{39.5} &\cellcolor[HTML]{DAE8FC} \textbf{63.12} &\cellcolor[HTML]{DAE8FC} 35.96 &\cellcolor[HTML]{DAE8FC} \textbf{59.28} &\cellcolor[HTML]{DAE8FC} 57.71 &\cellcolor[HTML]{DAE8FC} \textbf{70.33} &\cellcolor[HTML]{DAE8FC} \textbf{91.30} &\cellcolor[HTML]{DAE8FC} \textbf{68.57} &\cellcolor[HTML]{DAE8FC} \textbf{77.96} &\cellcolor[HTML]{DAE8FC} \textbf{64.94} &\cellcolor[HTML]{DAE8FC} \textbf{96.12} &\cellcolor[HTML]{DAE8FC} \textbf{62.41} \\
 & $\pm\Delta$ &  {\color[HTML]{FF0000} -4.70} & {\color[HTML]{00B050} +2.15} & {\color[HTML]{00B050} +11.94} & {\color[HTML]{FF0000} -11.15} & {\color[HTML]{00B050} +3.99} & {\color[HTML]{FF0000} -1.24}  & {\color[HTML]{00B050} +9.41} &{\color[HTML]{00B050} +25.65}  & {\color[HTML]{00B050} +9.26} & {\color[HTML]{00B050} +9.34} & {\color[HTML]{00B050} +6.59} & {\color[HTML]{00B050} +15.51} & {\color[HTML]{00B050} +6.39} \\
\midrule
CLIP & \multirow{2}{*}{ViT-L-14} & \textbf{40.23} & 47.94 & 58.33 & \textbf{60.21} & 64.89 & \textbf{73.78} & 69.23 & 69.88 & 72.15 & 76.83 & 67.11 & 87.5 & 65.67 \\
RemoteCLIP &  & \cellcolor[HTML]{DAE8FC} 37.22 &\cellcolor[HTML]{DAE8FC} \textbf{52.82} &\cellcolor[HTML]{DAE8FC} \textbf{70.83} &\cellcolor[HTML]{DAE8FC} 59.94 &\cellcolor[HTML]{DAE8FC} \textbf{66.32} &\cellcolor[HTML]{DAE8FC} 68.75 &\cellcolor[HTML]{DAE8FC} \textbf{79.84} &\cellcolor[HTML]{DAE8FC} \textbf{87.90} &\cellcolor[HTML]{DAE8FC} \textbf{72.32} &\cellcolor[HTML]{DAE8FC} \textbf{90.05} &\cellcolor[HTML]{DAE8FC} \textbf{74.90} &\cellcolor[HTML]{DAE8FC} \textbf{94.66} &\cellcolor[HTML]{DAE8FC} \textbf{71.30} \\
 &  $\pm\Delta$ &{\color[HTML]{FF0000} -3.01} & {\color[HTML]{00B050} +4.88} & {\color[HTML]{00B050} +12.50} & {\color[HTML]{FF0000} -0.27} & {\color[HTML]{00B050} +1.43} & {\color[HTML]{FF0000} -5.03} & {\color[HTML]{00B050} +10.61}& {\color[HTML]{00B050} +18.02} & {\color[HTML]{00B050} +0.17} & {\color[HTML]{00B050} +13.22} & {\color[HTML]{00B050} +7.79} & {\color[HTML]{00B050} +7.16} & {\color[HTML]{00B050} +5.63} \\
\bottomrule 
\end{tabular}%
}
\end{table*}

This section presents the zero-shot image classification outcome of our RemoteCLIP model. We used a total of 12 downstream datasets, including PatternNet~\cite{Zhou2017PatternNetAB}, EuroSAT~\cite{Helber2017EuroSATAN}, OPTIMAL-31~\cite{Wang2019SceneCW}, RSC11~\cite{Zhao2016FeatureSM}, AID~\cite{Xia2016AIDAB}, MLRSNet~\cite{Qi2020MLRSNetAM}, RSI-CB128~\cite{Li2017RSICBAL}, RSI-CB256~\cite{Li2017RSICBAL}, RESISC45~\cite{Cheng2017RemoteSI}, WHU-earth~\cite{Zhao2016DirichletDerivedMT}, WHU-RS19~\cite{Xia2010STRUCTURALHS}, and RS2800~\cite{Zou2015DeepLB}. We use the standard template-based prompting method, \textit{e.g.,} using ``\texttt{a satellite photo of \{class name\}.}'' as the text input to obtain the zero-shot classifier.

The details of evaluation results are shown in table~\ref{tab: zero-shot}. It can be seen from the experimental results that it has an overall improvement compared with the CLIP baseline. Overall, RemoteCLIP improves the averaged zero-shot accuracy improvement of +2.85\%, +6.39\%, and +5.63\% on 12 downstream datasets with three backbones, respectively. Our largest RemoteCLIP, the ViT-Large-14-based model, outperforms the CLIP counterpart on 9 out of 12 (75\%) datasets.  

However, the zero-shot performance of RemoteCLIP is consistently inferior to CLIP in some datasets. We suspect it is caused by the domain gap in image distribution. Our RemoteCLIP models are trained on a collection of high-resolution images (see Table~\ref{tab:dataset_detail} for detailed statics), but several downstream datasets, such as the EuroSAT dataset, have a much lower image resolution (\textit{e.g.,} 64 × 64). In addition, samples used to train RemoteCLIP usually covers rich semantics and variations, while several land cover classification datasets have a much different distribution (see visualizations in Fig.~\ref{fig:tsne}).

\subsubsection{Few-shot Classification}

Although the zero-shot classification performance of RemoteCLIP outperforms CLIP by a significant margin, in some datasets the accuracy is still far from satisfactory. In this section, we validate whether RemoteCLIP can be adapted to certain datasets with a few available training samples. We randomly sample few-shot training sets with 1, 4, 8, 16, and 32 shot samples, and use them to train an additional linear layer on top of the image representation via logistic regression. For logistic regression, the learning rate is set to 0.8, with SGD as the optimizer, and the CosineAnnealingLR scheduling strategy is used to automatically update the learning rate. We use CrossEntropyLoss as the criterion. We set the weight decay at 4e-5, the total number of epochs at 1000, and fix the batch size at 10,000. To choose suitable parameters for few-shot classification, we train the model through 5 iterations of a random search for optimal hyperparameters. Each iteration involves the use of distinct learning rates and weight decay coefficients while recording the accuracy achieved at each stage. Finally, the parameters associated with the best accuracy are used for the few-shot classification.

% We performed the 1,4,8,16, and 32-shot experiments on the twelve classification datasets described above, the results are shown on figure~\ref{fig:fewshot}. It is obvious that the RemoteCLIP method with ViT-B-32 has the highest accuracy on eleven other datasets except for WHU-earth. In addition, the RemoteCLIP method with ResNet-50 also performs better than most methods.

\begin{figure*}[h]
    \centering
	\begin{minipage}{0.24\linewidth}
		\centering
            \includegraphics[width=1\linewidth]{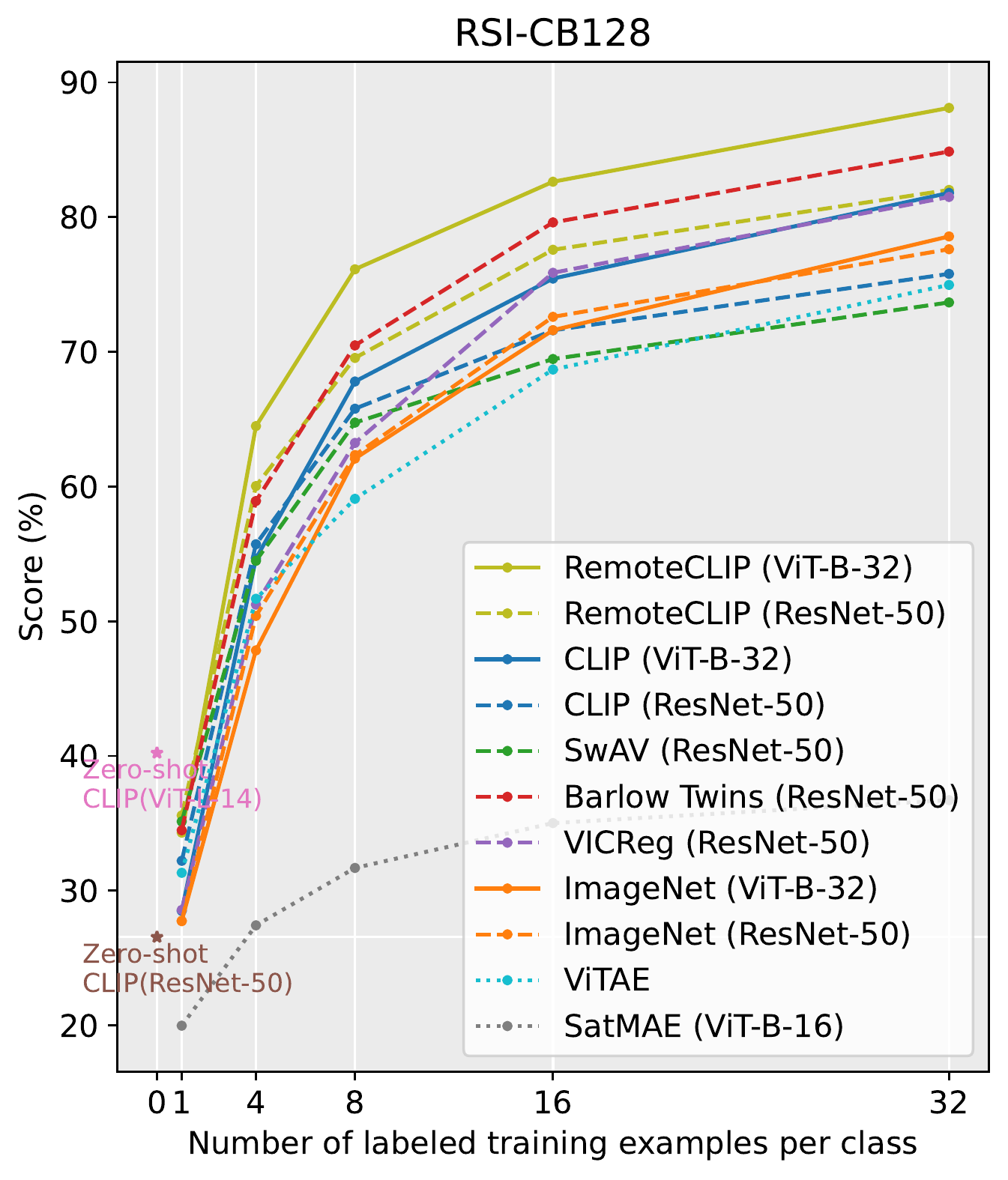}
	\end{minipage}
        \begin{minipage}{0.24\linewidth}
		\centering
		  \includegraphics[width=1\linewidth]{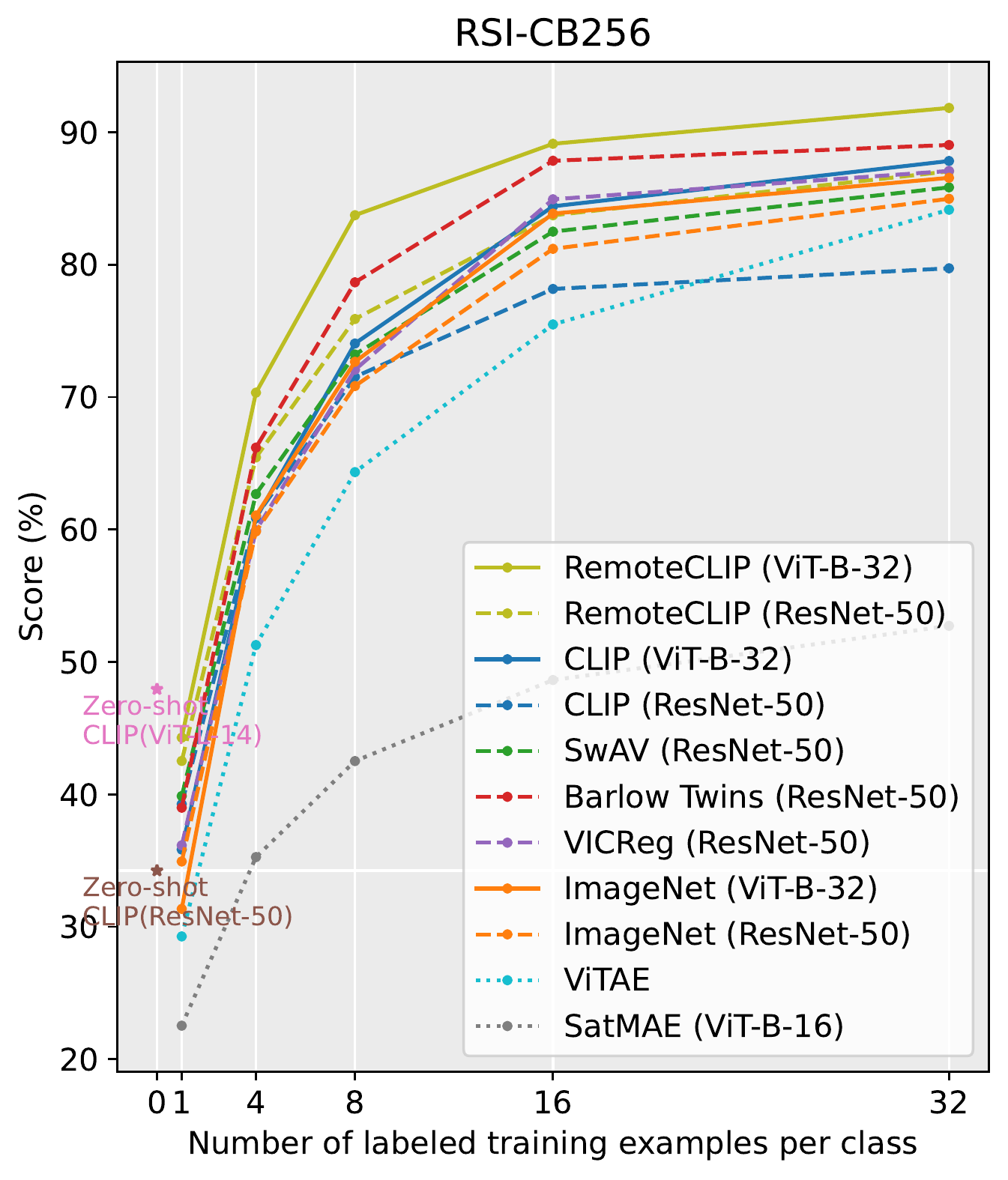}
	\end{minipage}
        \begin{minipage}{0.24\linewidth}
		\centering
		\includegraphics[width=1\linewidth]{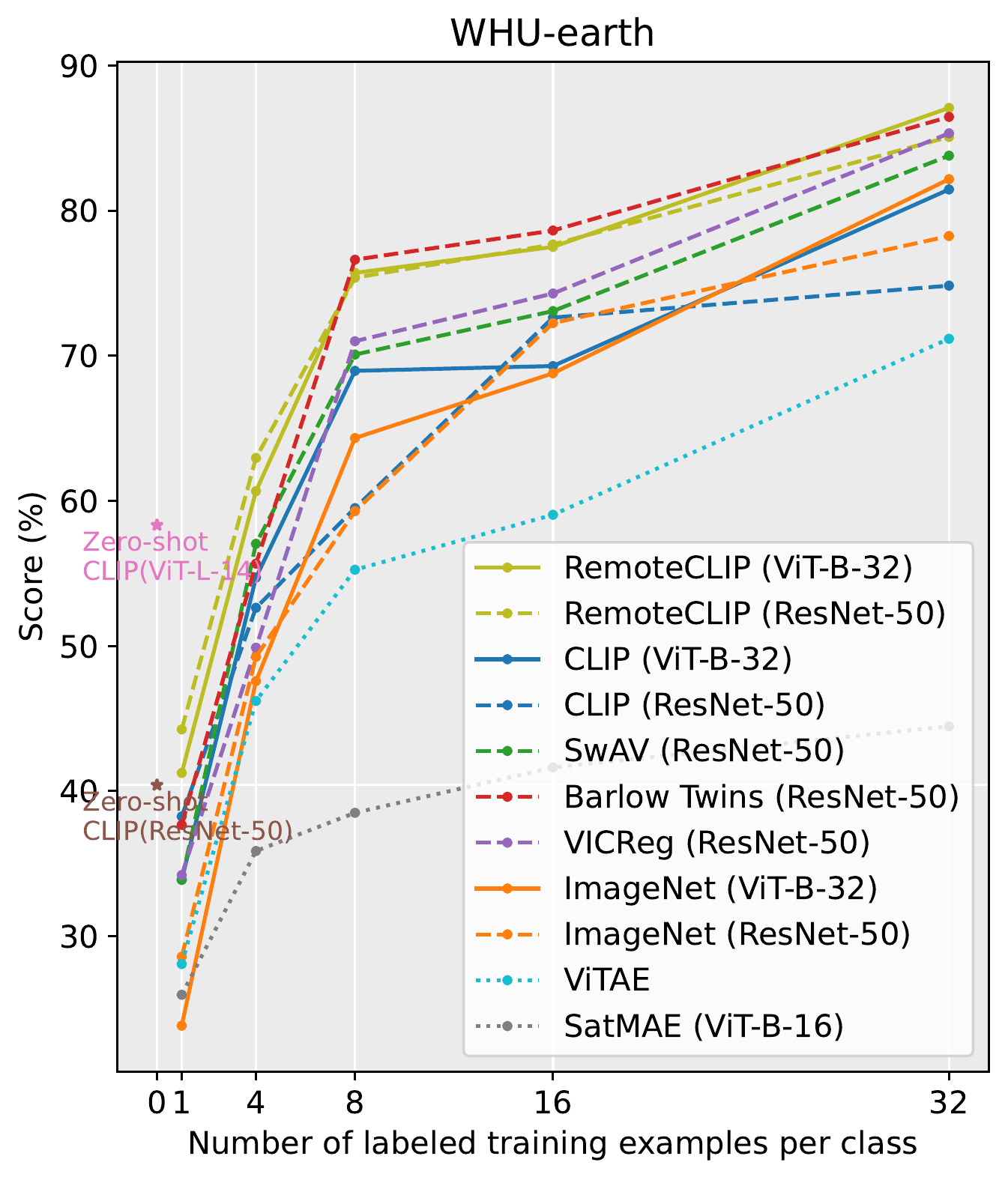}
	\end{minipage}
        \begin{minipage}{0.24\linewidth}
		\centering
		\includegraphics[width=1\linewidth]{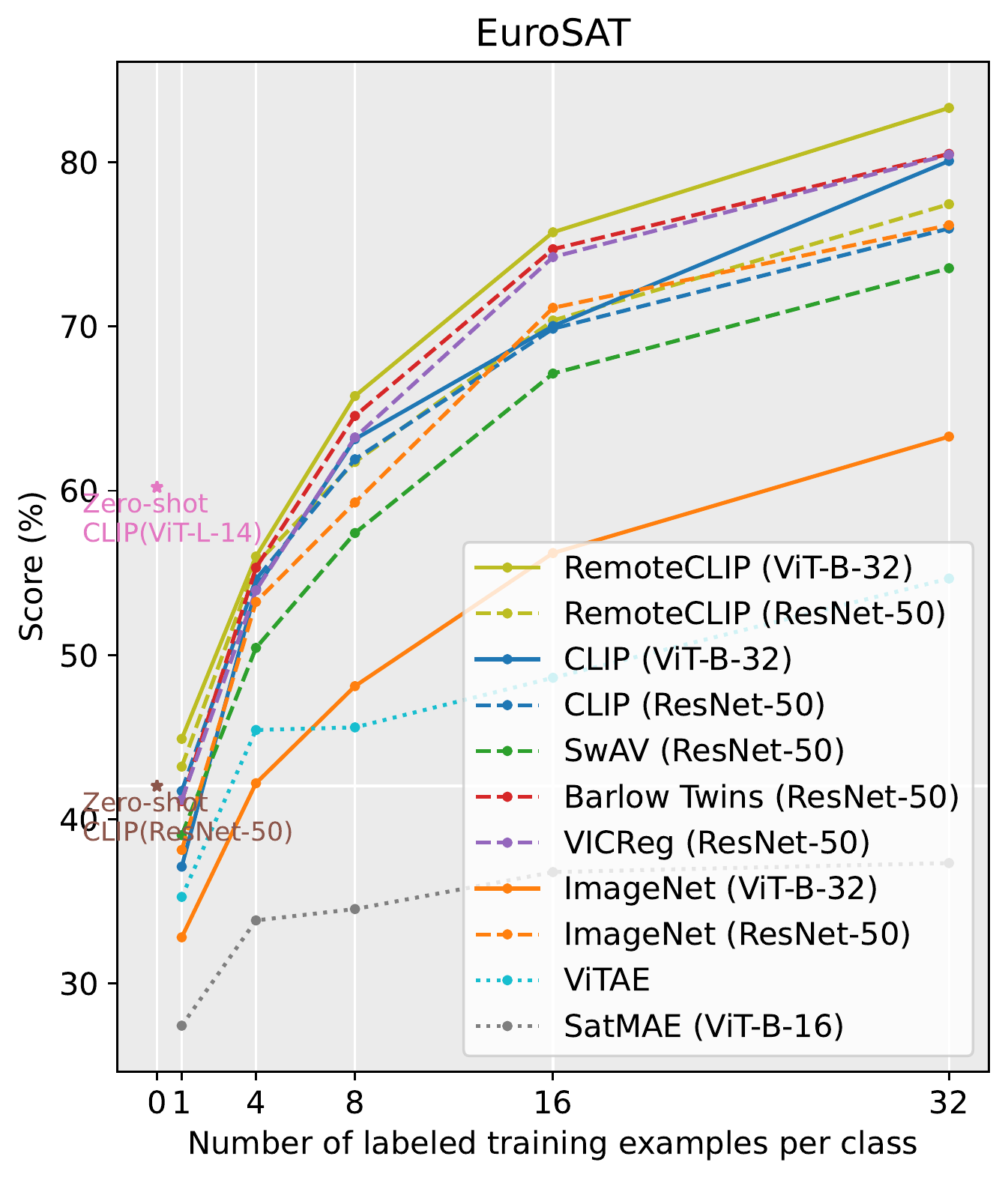}
	\end{minipage}

        \begin{minipage}{0.24\linewidth}
		\centering
		\includegraphics[width=1\linewidth]{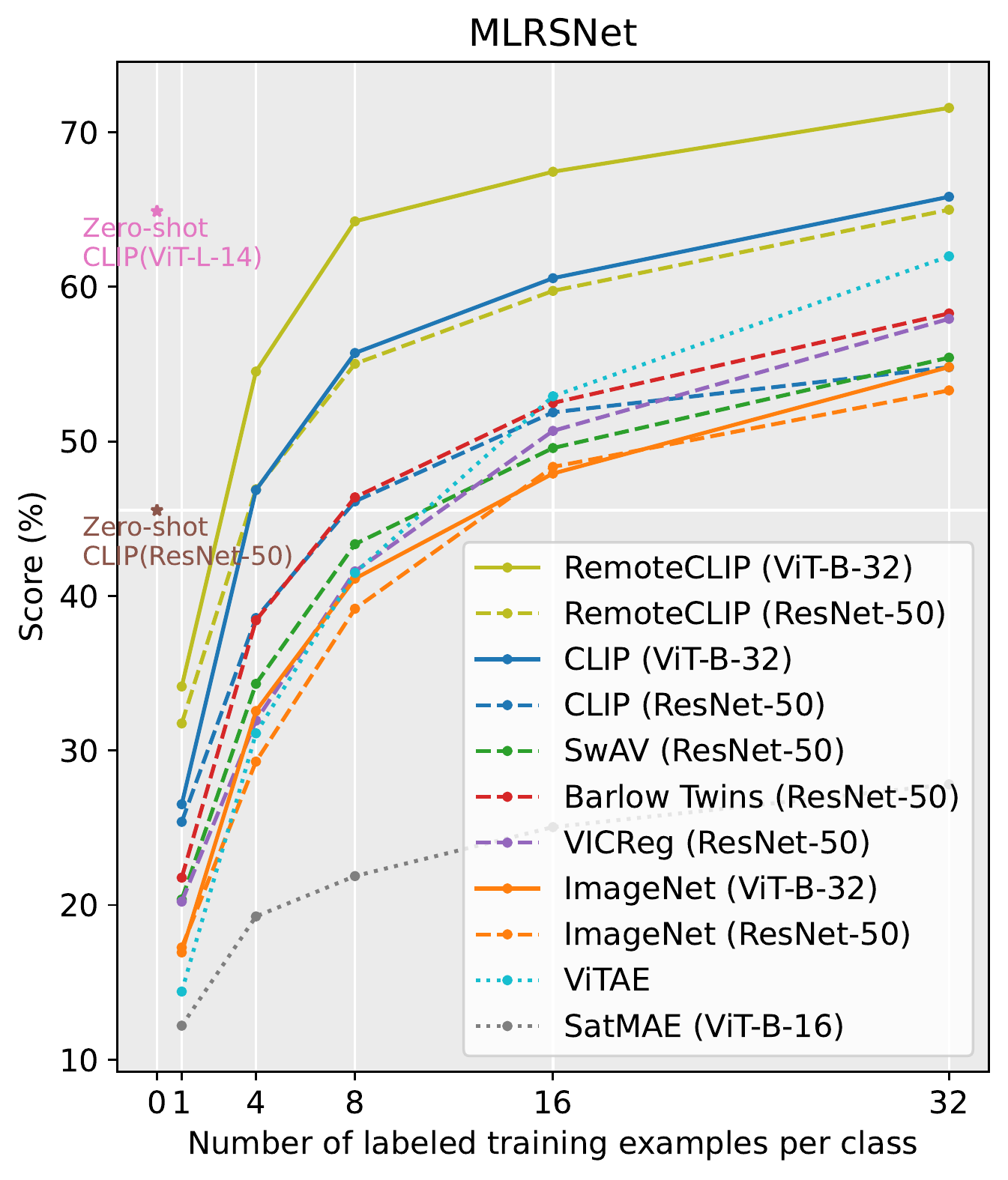}
	\end{minipage}
        \begin{minipage}{0.24\linewidth}
		\centering
		\includegraphics[width=1\linewidth]{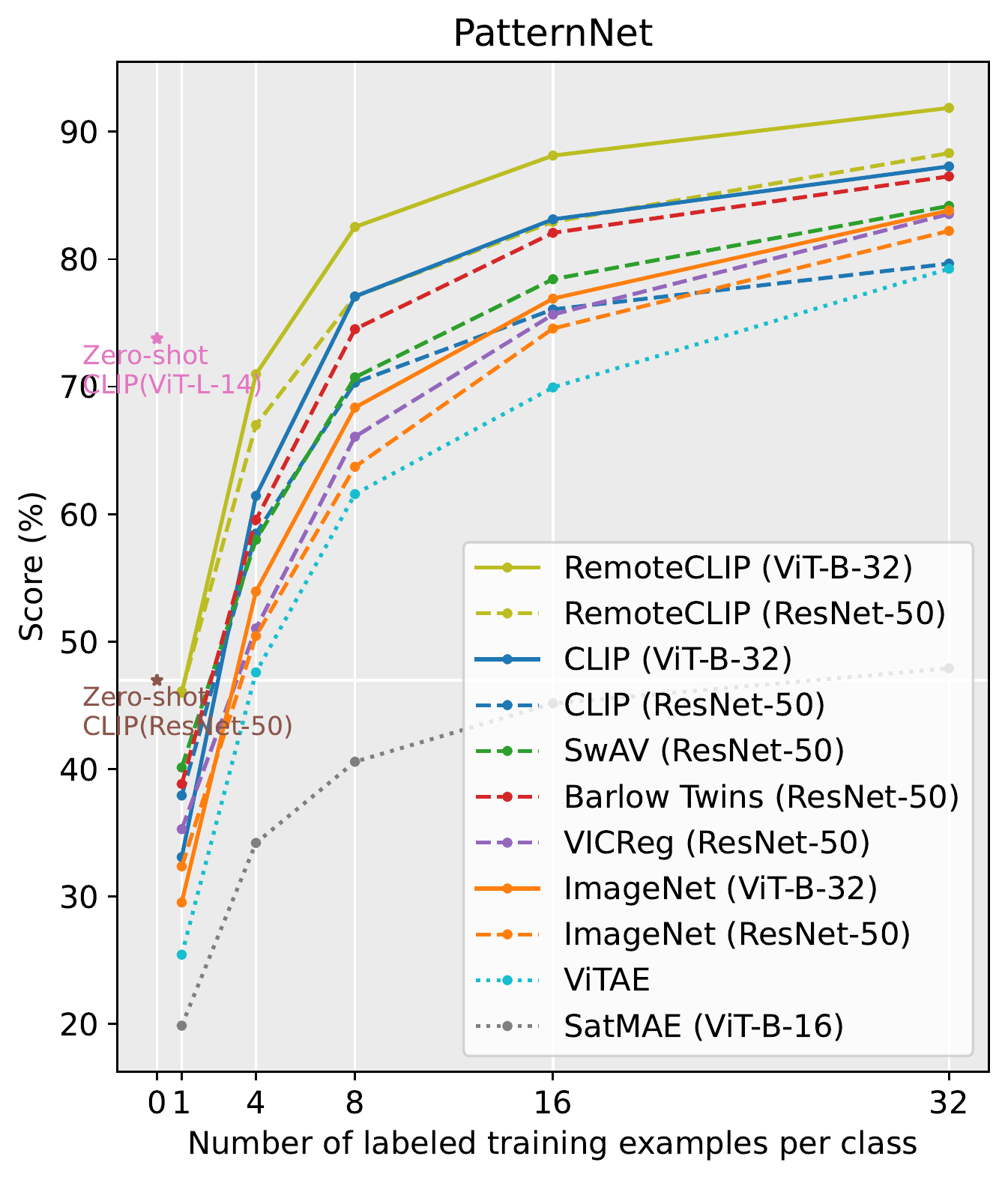}
	\end{minipage}
        \begin{minipage}{0.24\linewidth}
		\centering
		\includegraphics[width=1\linewidth]{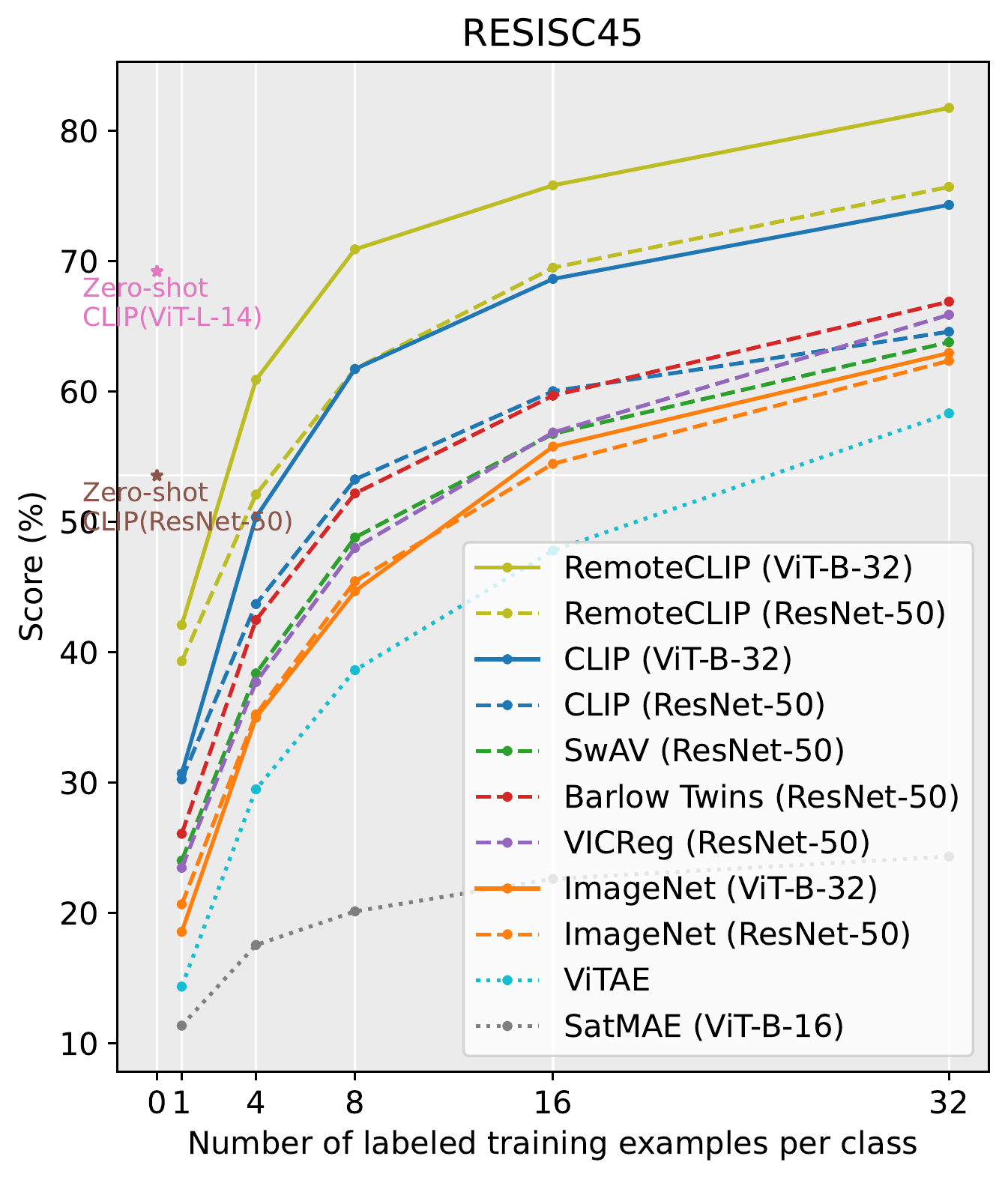}
	\end{minipage}
        \begin{minipage}{0.24\linewidth}
		\centering
		\includegraphics[width=1\linewidth]{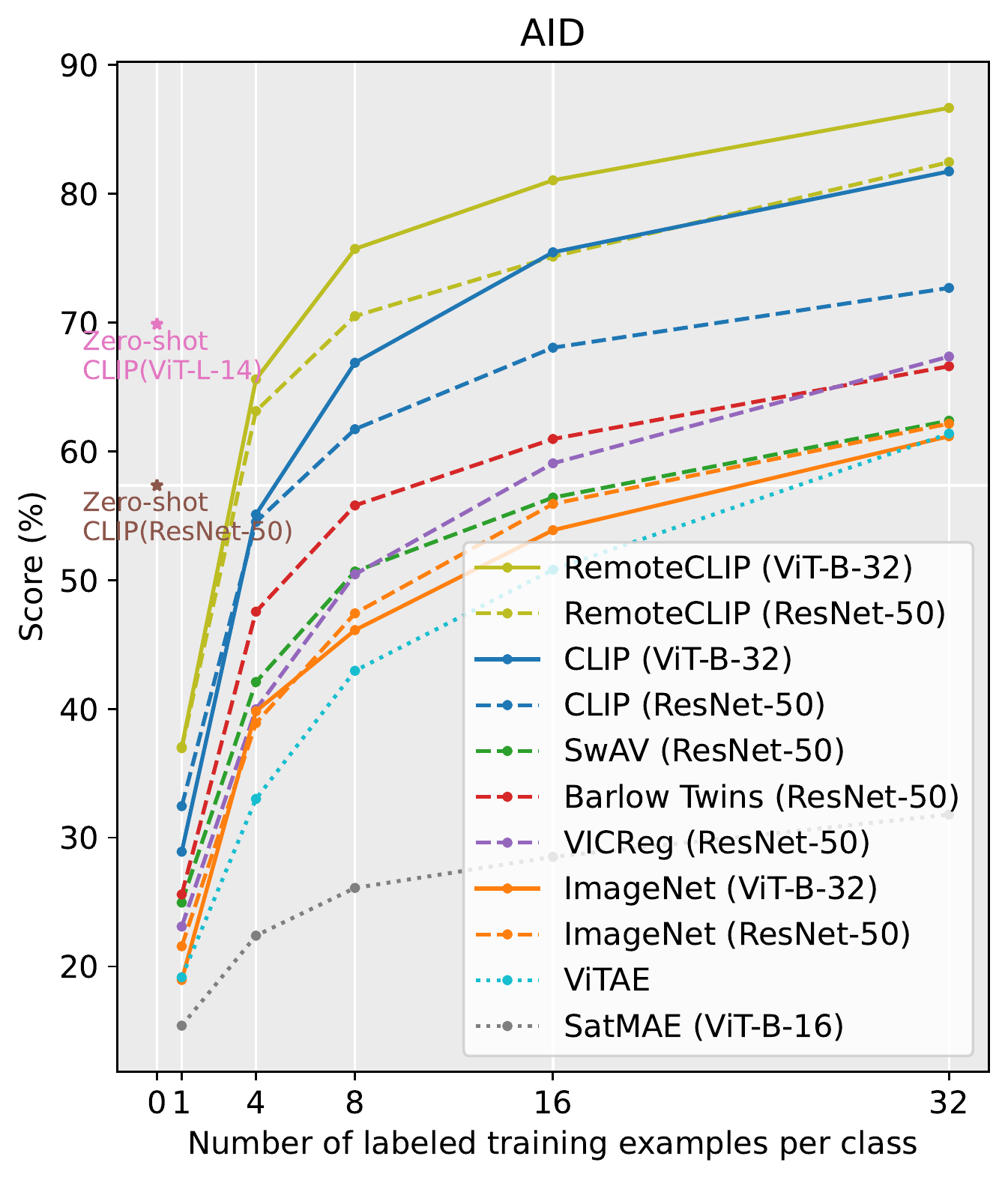}
	\end{minipage}

        \begin{minipage}{0.24\linewidth}
		\centering
		\includegraphics[width=1\linewidth]{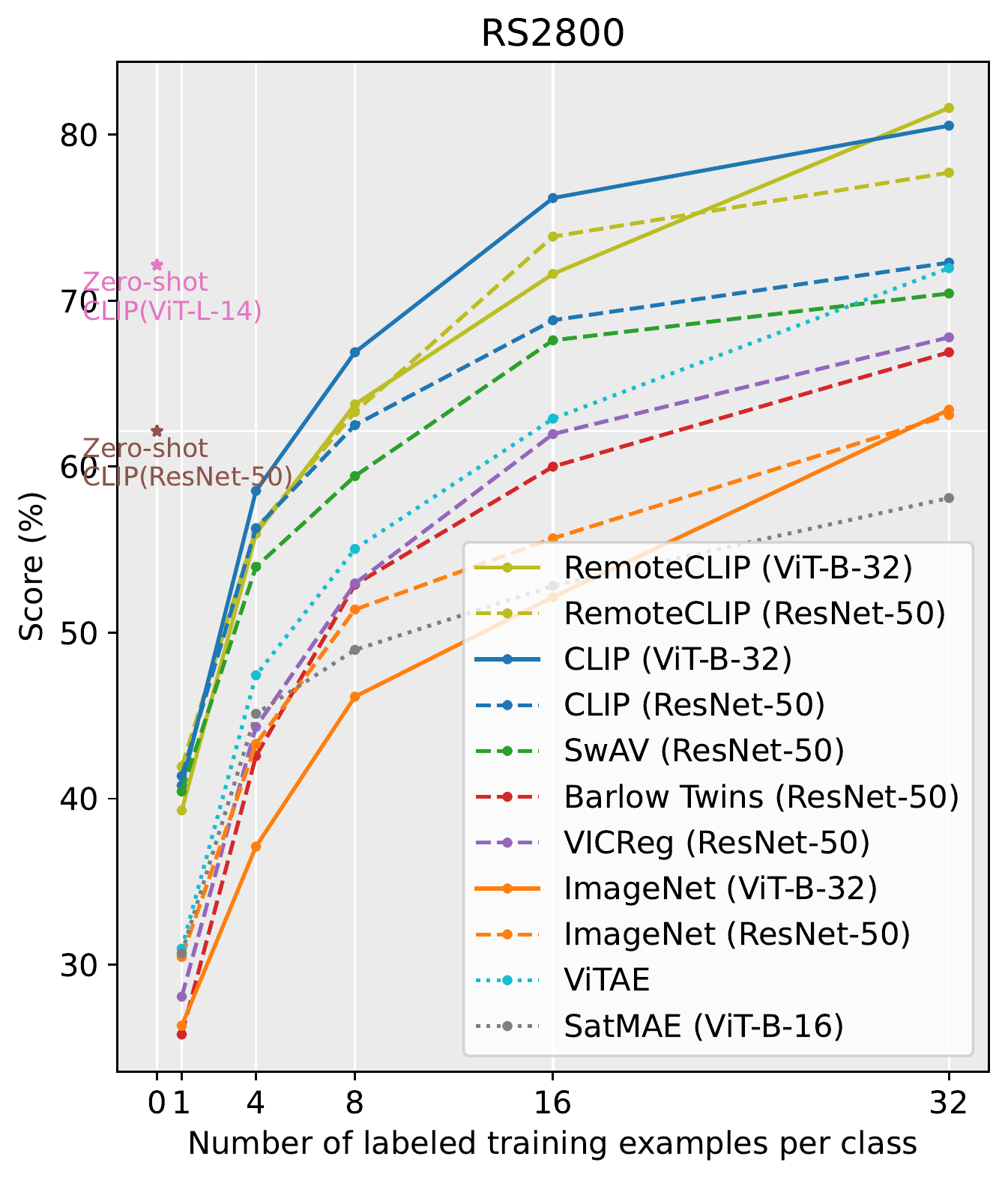}
	\end{minipage}
        \begin{minipage}{0.24\linewidth}
		\centering
		\includegraphics[width=1\linewidth]{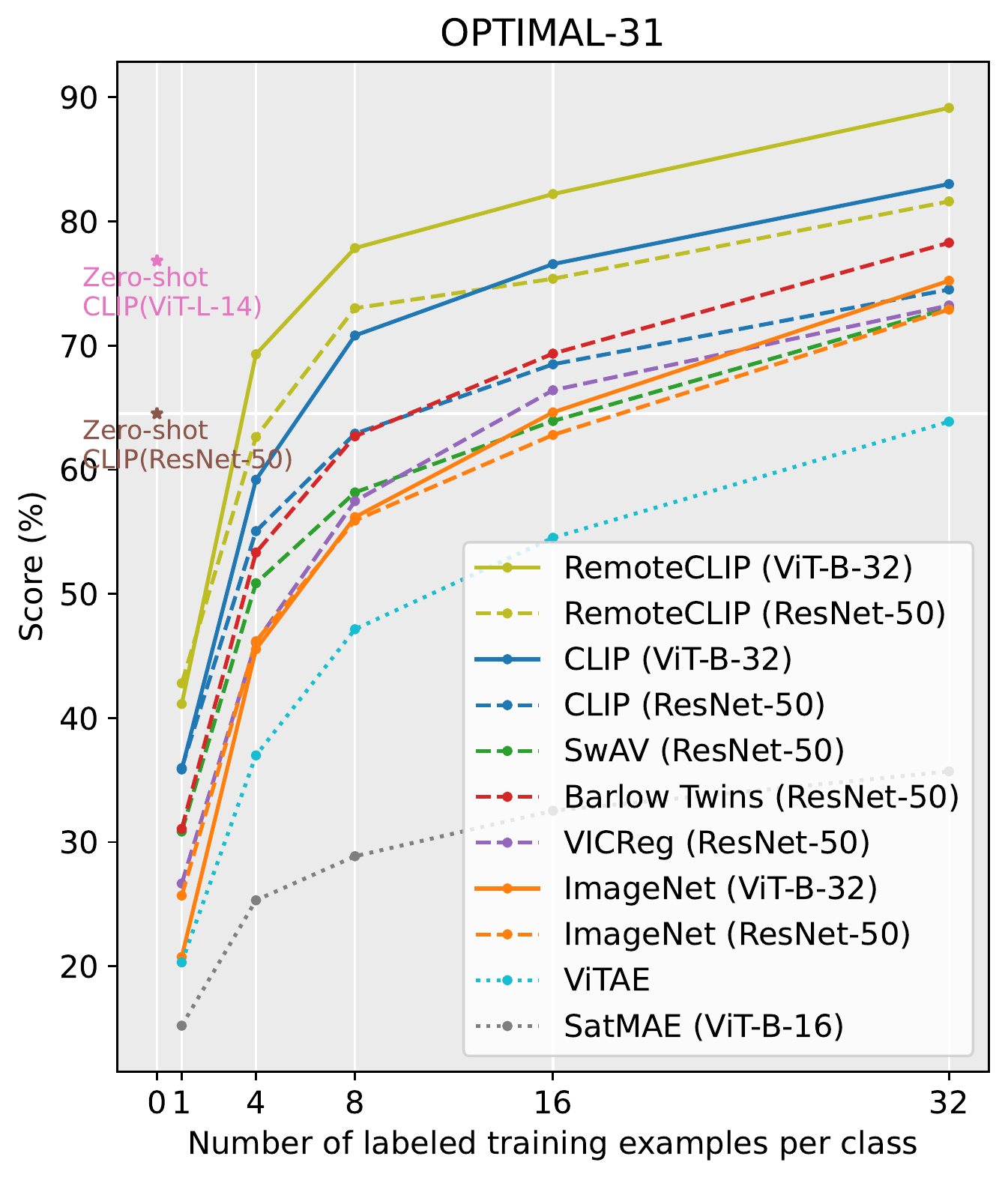}
	\end{minipage}
        \begin{minipage}{0.24\linewidth}
		\centering
		\includegraphics[width=1\linewidth]{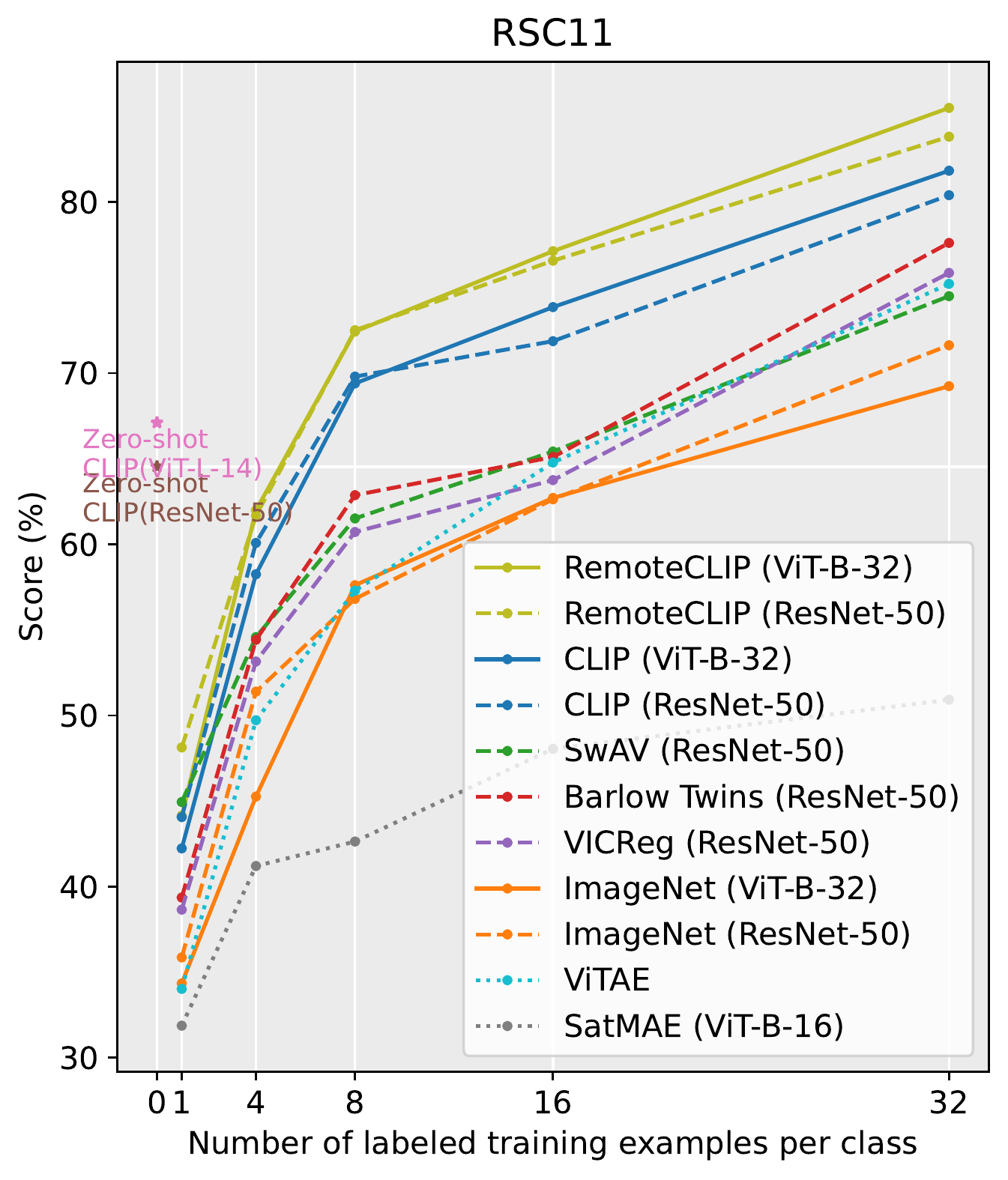}
	\end{minipage}
        \begin{minipage}{0.24\linewidth}
		\centering
		\includegraphics[width=1\linewidth]{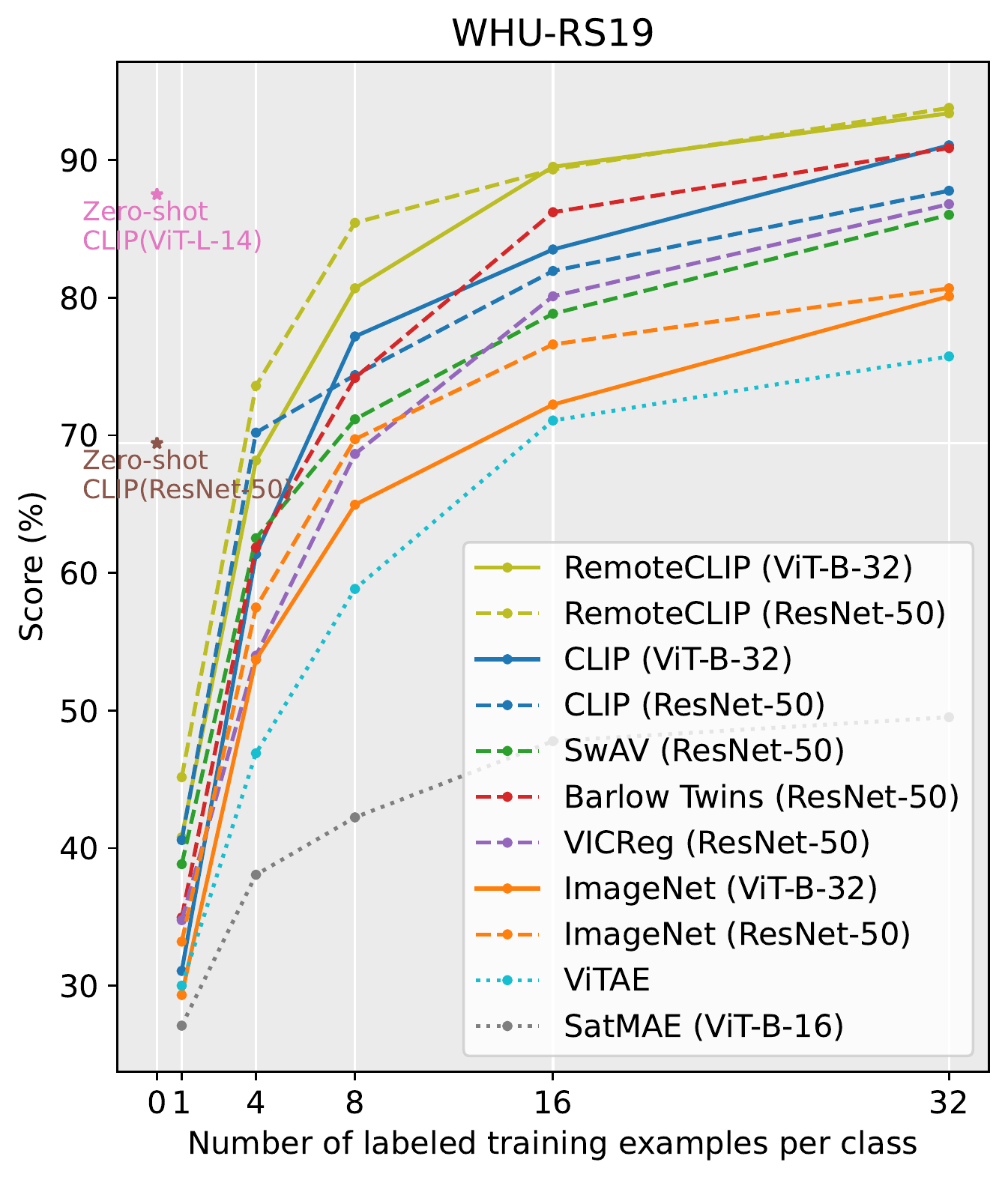}
	\end{minipage}

    \caption{Few-shot classification results on 12 remote sensing datasets. Zero-shot results of CLIP models are marked by \textcolor{brown}{brown} for ResNet-50 and by \textcolor{pink}{pink} for ViT-L-14. On the 32-shot setting, RemoteCLIP outperforms all compared models on all 12 datasets.}
    \label{fig:fewshot}
\end{figure*}

Fig.~\ref{fig:fewshot} shows the few-shot evaluation on 12 remote sensing classification datasets. We compare RemoteCLIP with a variety of baselines, including the vanilla CLIP model (ViT-Base-32 and ResNet-50), Self-supervised Learning (SSL-based) foundation visual models (SwAV, Barlow Twins, VICReg), ImageNet pretrained models (ViT-Base-32 and ResNet-50), and existing remote sensing foundation models (ViTAE and SatMAE). Visualization of experimental results shows that a few-shot training set could significantly boost the performance of RemoteCLIP models in all datasets. Using 32-shot samples, the RemoteCLIP model outperforms all compared baselines in all 12 datasets.

\subsubsection{Full-shot Linear Probing and \textit{k}-NN Classification}

Finally, we turn to benchmark RemoteCLIP for conventional linear probing (linear classification) and \textit{k}-NN classification. We use the same 12 classification datasets previously used for zero-shot and few-shot evaluation. For linear classification, the hyperparameter settings are the same as those for the few-shot classification experiment. For k-NN classification, the number of nearest neighbors k is set to 20 and the temperature parameter T is set to 0.07. We take the accuracy of the top 1 category as the output of k-NN classification.

The results are shown in table~\ref{tab: linear-probe and K-NN}. We find that the classification performance of RemoteCLIP is better than CLIP and other self-supervised models. It is not surprising that RemoteCLIP produces such strong visual representations. As mentioned in Section~\ref{sec:clip}, the vanilla CLIP model can already outperform a variety of foundation visual models in linear probing on remote sensing datasets. RemoteCLIP further enhances such representation.

\begin{table*}[]
\centering
\caption{Linear probing and and \textit{k}-NN classification results on 12 remote sensing datasets.}
\label{tab:  linear-probe and K-NN}
\resizebox{\textwidth}{!}{%
\begin{tabular}{cccccccccccccccccccccccccccc}
\toprule
\multirow{2}{*}{\textbf{Method}} & \multirow{2}{*}{\textbf{Backbone}} & \multicolumn{2}{c}{\textbf{RSI-CB128}} & \multicolumn{2}{c}{\textbf{RSI-CB256}} & \multicolumn{2}{c}{\textbf{WHU-earth}} & \multicolumn{2}{c}{\textbf{EuroSAT}} & \multicolumn{2}{c}{\textbf{MLRSNet}} & \multicolumn{2}{c}{\textbf{PatternNet}} & \multicolumn{2}{c}{\textbf{RESISC45}} & \multicolumn{2}{c}{\textbf{AID}} & \multicolumn{2}{c}{\textbf{RS2800}} & \multicolumn{2}{c}{\textbf{OPTIMAL-31}} & \multicolumn{2}{c}{\textbf{RSC11}} & \multicolumn{2}{c}{\textbf{WHU-RS19}} & \multicolumn{2}{c}{\textbf{Average}} \\
\cline{3-28}
 &  & \textbf{Linear} & \textbf{\textit{k}-NN} & \textbf{Linear} & \textbf{\textit{k}-NN} & \textbf{Linear} & \textbf{\textit{k}-NN} & \textbf{Linear} & \textbf{\textit{k}-NN} & \textbf{Linear} & \textbf{\textit{k}-NN} & \textbf{Linear} & \textbf{\textit{k}-NN} & \textbf{Linear} & \textbf{\textit{k}-NN} & \textbf{Linear} & \textbf{\textit{k}-NN} & \textbf{Linear} & \textbf{\textit{k}-NN} & \textbf{Linear} & \textbf{\textit{k}-NN} & \textbf{Linear} & \textbf{\textit{k}-NN} & \textbf{Linear} & \textbf{\textit{k}-NN} & \textbf{Linear} & \textbf{\textit{k}-NN} \\
 \midrule

ImageNet &  \multirow{7}{*}{ResNet-50}  & 95.69 & 93.24 & 97.92 & 97.40 & 92.92 & 93.69 & 91.48 & 88.41 & 78.98 & 74.78 & 96.18 & 93.45 & 86.16 & 83.60 & 83.00 & 79.45 & 75.89 & 79.29 & 87.10 & 86.29 & 79.68 & 78.09 & 95.63 & 90.21 & 87.32 & 87.08 \\
SwAV &   & 95.27 & 95.61 & 98.59 & 98.17 & 95.20 & 93.96 & 91.17 & 91.37 & 79.04 & 76.12 & 96.94 & 94.18 & 88.60 & 85.59 & 86.00 & 80.80 & 81.07 & 86.07 & 88.44 & 84.14 & 84.86 & 78.89 & 96.12 & 92.23 & 88.97 & 88.83 \\
Barlow Twins &   & 98.07 & 95.91 & 99.03 & 98.13 & 95.83 & 95.42 & 94.78 & 91.57 & 82.41 & 77.55 & 97.73 & 93.83 & 91.10 & 86.10 & 88.25 & 81.75 & 77.32 & 86.07 & 91.94 & 86.83 & 85.26 & 78.09 & 97.09 & 91.75 & 90.00 & 89.73 \\
VICReg &   & 97.47 & 96.03 & 98.67 & 98.21 & 95.21 & 94.79 & 95.06 & 91.44 & 82.59 & 78.02 & 96.83 & 94.03 & 91.03 & 86.75 & 88.10 & 81.50 & 77.86 & 86.79 & 90.59 & 86.83 & 84.46 & 77.69 & 96.60 & 90.78 & 89.85 & 89.56 \\
CLIP &  & 94.89 & 94.05 & 97.30 & 97.24 & 93.12 & 91.88 & 91.67 & 88.54 & 80.08 & 77.14 & 95.61 & 92.86 & 85.73 & 85.65 & 90.95 & 86.90 & 83.75 & 81.43 & 88.99 & 87.63 & 87.65 & 87.25 & 97.57 & 93.69 & 89.47 & 89.42 \\
CLIP-CL &   & 95.99 & 94.92 & 98.41 & 98.09 & 96.25 & 94.79 & 89.80 & 87.65 & 79.32 & 76.99 & 97.30 & 95.15 & 89.10 & 88.19 & 94.80 & 92.85 & 82.50 & 89.29 & 91.40 & 89.78 & 91.63 & 84.86 & 98.06 & 97.57 & 91.18 & 91.25 \\
 \rowcolor[HTML]{DAE8FC}RemoteCLIP &   &  96.06 & 94.78 & 98.39 & 97.62 & 95.42 & 95.63 & 92.56 & 90.20 & 83.32 & 81.21 & 97.37 & 95.95 & 90.94 & 90.05 & 94.35 & 90.10 & 85.00 & 89.46 & 92.74 & 90.86 & 91.63 & 85.66 & 98.06 & 95.63 & 92.06 & 92.04 \\
\midrule
ImageNet &   \multirow{4}{*}{ViT-Base} & 96.45 & 91.29 & 98.11 & 97.00 & 93.75 & 91.67 & 85.57 & 76.56 & 78.61 & 74.05 & 96.81 & 92.98 & 86.89 & 81.63 & 83.55 & 76.45 & 78.93 & 78.04 & 89.51 & 81.18 & 81.67 & 80.88 & 94.17 & 89.81 & 86.34 & 86.05 \\
ViTAE & & 93.10 & 95.65 & 98.41 & 94.05 & 93.33 & 78.96 & 61.41 & 82.27 & 91.15 & 80.37 & 98.50 & 90.82 & 87.94 & 65.33 & 88.30 & 64.05 & 92.86 & 78.93 & 86.29 & 54.84 & 92.83 & 71.31 & 91.74 & 70.39 & 84.02 & 83.03 \\
CLIP & & 97.36 & 94.17 & 98.55 & 97.40 & 95.00 & 92.08 & 95.15 & 90.28 & 85.43 & 82.26 & 97.58 & 94.36 & 92.60 & 89.73 & 94.95 & 90.35 & 88.57 & 88.21 & 93.55 & 90.86 & 90.84 & 86.85 & 97.09 & 93.69 & 92.31 & 92.15 \\
\rowcolor[HTML]{DAE8FC}RemoteCLIP &  &   98.02 & 95.82 & 99.01 & 98.51 & 95.42 & 97.08 & 96.19 & 93.50 & 87.00 & 85.11 & 98.47 & 97.32 & 94.27 & 92.67 & 95.95 & 92.55 & 86.96 & 87.86 & 95.97 & 94.35 & 91.63 & 89.24 & 97.57 & 94.17 & 93.93 & 93.77 \\
% \midrule
% SatMAE & ViT-B-16 & 55.56 & 74.56 & 75.70 & 81.16 & 54.38 & 60.00 & 37.46 & 62.83 & 44.09 & 50.92 & 65.36 & 69.04 & 35.22 & 37.33 & 39.90 & 40.00 & 68.04 & 61.96 & 40.59 & 34.13 & 57.37 & 54.58 & 58.74 & 50.97 & 54.74 & 54.54 \\
% \midrule
% ViTAE &  & 93.90 & 95.27 & 98.37 & 94.26 & 93.54 & 81.67 & 72.04 & 87.54 & 89.18 & 79.61 & 98.37 & 90.36 & 84.54 & 63.92 & 87.40 & 63.75 & 91.43 & 78.04 & 87.37 & 51.34 & 92.43 & 70.52 & 91.26 & 66.02 & 84.18 & 82.97 \\
% \midrule

\bottomrule
\end{tabular}%
}
\end{table*}

\subsection{Ablation Study}

% \textcolor{red}{TODO: contents and Tables need revisions @ Zhou Xiao Cong}

\textbf{Backbone ablation}: In Table~\ref{tab: Backbone ablation}, we investigate the effects of the image and text backbones through ablation experiments conducted on the \texttt{Ret-3 + Det-10 + Seg-4} dataset using RemoteCLIP. The results indicate that the optimal outcome is achieved when both the image and text backbones are pre-trained. Furthermore, the experiment highlights the greater significance of pre-training the image backbone compared to pre-training the text backbone.

\textbf{Pre-training model ablation}: The experimental findings can be observed in Table~\ref{tab: Pre-training model ablation}. When comparing RemoteCLIP to prior pre-training techniques, notable advancements are observed in both the Retrieval task and Zero-shot tasks. Specifically, RemoteCLIP exhibits substantial improvements of approximately 10\% and 15\% in the Retrieval task and Zero-shot tasks, respectively.
 
\textbf{Dataset ablation}: To explore the validity of sentence-making rules, we conduct an ablation experiment on the dataset using RemoteCLIP. We apply different sentence-making rules and evaluated their impact on the same test set. The experimental results, presented in Table~\ref{tab: Dateset ablation}, reveal superior performance with the \texttt{Ret-3 + Det-10 + Seg-4} dataset. These findings indicate the task's demand for richer textual information and affirm the effectiveness of our sentence-making strategy.

\textbf{Preprocessing ablation}: To ensure controlled conditions, we conduct ablation experiments on the \texttt{RET-3} dataset. As shown in Table~\ref{tab: Preprocessing ablation}, the retrieval results exhibit higher performance with the application of rotation.

\textbf{Loss Ablation}: \textcolor{black}{To validate the superiority of the InfoNCE loss for RemoteCLIP, we conducted experiments comparing various loss functions. Our findings, as depicted in Table ~\ref{tab: Loss ablation}, demonstrate that InfoNCE effectively captures the semantic correlation between images and texts. It excels in distinguishing similarities and differences among samples, thereby leading to more robust feature representations. As a result, our experiments show that InfoNCE achieves the most favorable results, as evidenced by its superior performance across both retrieval average and zero-shot average.}

\begin{figure}
    \centering
    \includegraphics[width=\linewidth]{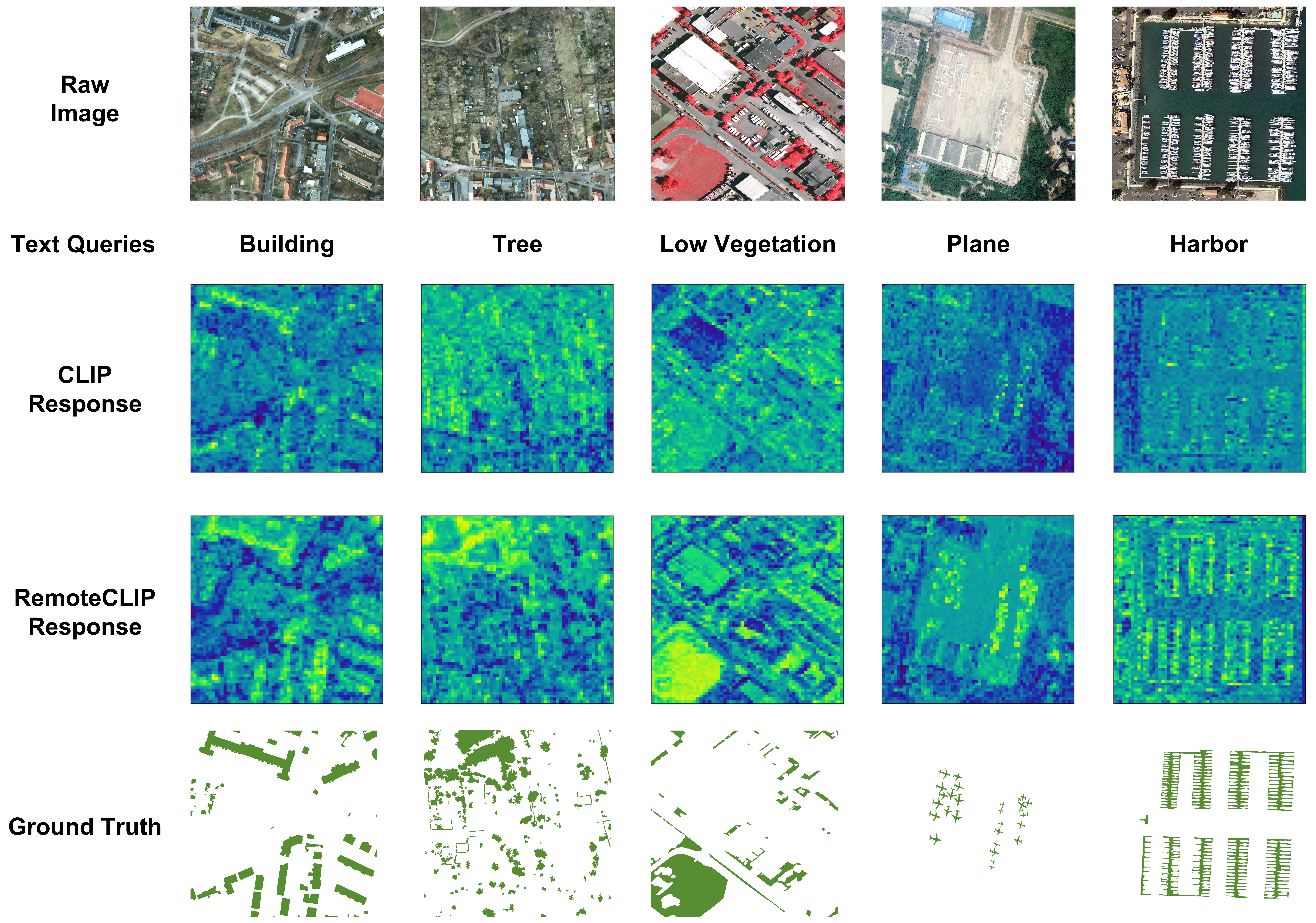}
    \caption{CLIP vs RemoteCLIP visualization of similarity across different categories. Top row: Raw images of different datasets. Above the top row are the categories for which similarity is to be calculated. Second row: Ground truth masks. For convenience, we use the same color to represent their interested category. Third row: Visualization of image-text similarity calculated by CLIP. Bottom row: Visualization of image-text similarity calculated by RemoteCLIP.}
    \label{fig:similarity_map_v2}
\end{figure}

\begin{table*}[!h]
    % \centering
    \caption{Ablation study results. The first row from left to right: pretraining ablation, backbone ablation and dataset ablation. The second row from left to right: preprocessing ablation and loss ablation. Settings adopted by RemoteCLIP are marked in blue.}
    \label{tab:ablation}

    % \usepackage{multirow}
    % \usepackage{graphicx}
    % [Backbone ablation]

    % \subtable{Dateset ablation}

    % [Pre-training model]
    % \subfloat[Pre-training model ablation]
    \hfill
    {
    \centering
     \label{tab: Pre-training model ablation}
    \resizebox{0.32\linewidth}{!}{%
    \begin{tabular}{cccc}
    \textbf{Backbone} & \textbf{Method} & \textbf{\begin{tabular}[c]{@{}c@{}}Retrival\\ Average\end{tabular}} & \textbf{\begin{tabular}[c]{@{}c@{}}Zero-shot\\ Average\end{tabular}} \\
    \toprule
    \multirow{5}{*}{ResNet-50} & ImageNet & 37.07 & 44.36 \\
     & SwAV & 34.6 & 44.59 \\
     & VICReg & 34.28 & 41.01 \\
     & BarlowTwins & 32.95 & 40.36 \\
     & \cellcolor[HTML]{DAE8FC} CLIP & \cellcolor[HTML]{DAE8FC}\textbf{42.01} & \cellcolor[HTML]{DAE8FC}\textbf{55.06} \\
     \midrule
    \multirow{5}{*}{ViT-Base} & ViTAE & 39.08 & 47.85 \\
     & ViTAE & 38.75 & 48.5 \\
     & DINOv2 & 38.14 & 50.24 \\
     & ImageNet & 35.08 & 46.19 \\
     & \cellcolor[HTML]{DAE8FC} CLIP & \cellcolor[HTML]{DAE8FC}\textbf{47.00} & \cellcolor[HTML]{DAE8FC}\textbf{64.52} \\
     \bottomrule
    \end{tabular}%
    }
    }
    \hfill
    % \subfloat[Backbone ablation]
    {
    \centering
    \label{tab: Backbone ablation}
    \resizebox{0.32\linewidth}{!}{%
    \begin{tabular}{ccccc}
    \textbf{} & \textbf{\begin{tabular}[c]{@{}c@{}}Image\\ Pre-trained\end{tabular}} & \textbf{\begin{tabular}[c]{@{}c@{}}Text\\ Pre-trained\end{tabular}} & \textbf{\begin{tabular}[c]{@{}c@{}}Retrival\\ Average\end{tabular}} & \textbf{\begin{tabular}[c]{@{}c@{}}Zero-shot\\ Average\end{tabular}} \\
    \toprule
    \multirow{4}{*}{ResNet-50} & \cellcolor[HTML]{DAE8FC} \checkmark & \cellcolor[HTML]{DAE8FC}\checkmark & \cellcolor[HTML]{DAE8FC}\textbf{42.01} & \cellcolor[HTML]{DAE8FC}\textbf{53.46} \\
     & $\times$ & \checkmark & 25.56 & 37.46 \\
     & \checkmark & $\times$ & 35.72 & 46.03 \\
     & $\times$ & $\times$ & 24.44 & 36.97 \\
     \midrule
    \multirow{4}{*}{ViT-B-32} & \cellcolor[HTML]{DAE8FC} \checkmark & \cellcolor[HTML]{DAE8FC}\checkmark & \cellcolor[HTML]{DAE8FC}\textbf{47.00} & \cellcolor[HTML]{DAE8FC}\textbf{64.52} \\
     & $\times$ & \checkmark & 21.56 & 42.60 \\
     & \checkmark & $\times$ & 37.13 & 54.30 \\
     & $\times$ & $\times$ & 18.92 & 30.93 \\
     \bottomrule
    \end{tabular}%
    } 
    }
    \hfill
    % Dateset ablation
    % \subfloat[Dateset ablation]
    {
    \centering
    \resizebox{0.32\linewidth}{!}{%
    \label{tab: Dateset ablation}
    \begin{tabular}{ccccc}
    \texttt{SEG-4} & \texttt{DET-10} & \texttt{RET-3} & \textbf{\begin{tabular}[c]{@{}c@{}}Retrival\\ Average\end{tabular}} & \textbf{\begin{tabular}[c]{@{}c@{}}Zero-shot\\ Average\end{tabular}} \\
    \toprule
    \checkmark & $\times$ & $\times$ & 7.15 & 14.55 \\
    $\times$ & \checkmark & $\times$ & 9.82 & 21.37 \\
    $\times$ & $\times$ & \checkmark & 36.32 & 48.75 \\
    \checkmark & \checkmark &  & 10.23 & 24.09 \\
    \checkmark & $\times$ & \checkmark & 37.24 & 46.94 \\
    $\times$ & \checkmark & \checkmark & 39.72 & 51.31 \\
    \rowcolor[HTML]{DAE8FC} \checkmark & \checkmark & \checkmark & \textbf{42.01} & \textbf{53.46} \\
    \bottomrule
    \end{tabular}%
    }
    } % \subcaption{Dateset ablation} 
    % \subcaption{Dateset ablation}
    
    \vspace{0.5cm}
    \hspace{3cm}
     % Preprocessing ablation
    % \subfloat[Preprocessing ablation]
    {
    \centering
    \label{tab: Preprocessing ablation}
    \resizebox{0.32\linewidth}{!}{%
    \begin{tabular}{ccc}
    \textbf{Preprocessing} & \textbf{\begin{tabular}[c]{@{}c@{}}Retrival\\ Average\end{tabular}} & \textbf{\begin{tabular}[c]{@{}c@{}}Zero-shot\\ Average\end{tabular}} \\
    \toprule
    \rowcolor[HTML]{DAE8FC} Rotation Augmentation & \textbf{38.90} & 47.98 \\
    No Augmentation & 37.74 & 48.05 \\
    Super Resolution & 37.98 & 47.18 \\
    SimCLR Augmentation & 37.99 & \textbf{48.07} \\
    \bottomrule
    \end{tabular}%
    }
    }
    \hspace{0.5cm}
     % Loss ablation
    % \subfloat[Loss ablation]
    {
    \centering
    \label{tab: Loss ablation}
    \resizebox{0.32\linewidth}{!}{%
    \begin{tabular}{ccc}
    \textbf{Loss} & \textbf{\begin{tabular}[c]{@{}c@{}}Retrival\\ Average\end{tabular}} & \textbf{\begin{tabular}[c]{@{}c@{}}Zero-shot\\ Average\end{tabular}} \\
    \toprule
    \rowcolor[HTML]{DAE8FC} InfoNCE
     & \textbf{36.32} & \textbf{48.57} \\
    % UniCL~\cite{Yang2022Unified} & 34.41 & 47.64 \\
    Margin Ranking~\cite{Cao2007Learning} & 28.93 & 48.47 \\
    SigLIP~\cite{Zhai2023Sigmoid} & 26.68 & 45.66 \\\
    N-pair~\cite{Sohn2016Improved} & 25.31 & 45.52 \\
    BarlowTwins~\cite{Zbontar2021Barlow} & 21.03 & 35.44 \\
    \bottomrule
    \end{tabular}%
    }
    }
    
\end{table*}

\subsection{Feature Visualization}

\textcolor{black}{To demonstrate that RemoteCLIP has learned richer remote sensing semantic information, we visualize the similarity scores between images and relevant categories. Specifically, we cropped high-resolution images (from Potsdam, Vaihingen, and iSAID dataset) into 64$\times$64=4096 patches with 1/3 overlap between near by patches. We employ ``\texttt{a \{target class name\}}'' as our text prompt, and calculate cosine similarity between patch visual feature and textual feature. We visualize the similarity score with respect to different categories in Fig.~\ref{fig:similarity_map_v2}. Compared to that of original CLIP, the feature similarity response of RemoteCLIP has a better correlation with the ground truth mask annotation. RemoteCLIP demonstrated the ability to roughly locate the spatial position of the target category which suggests that RemoteCLIP not only learns a rich semantic information but also holds promise for tasks related to remote sensing visual localization, such as remote sensing object detection etc.}

\textcolor{black}{}

\section{Conclusion}
\label{sec:conclusion}

In this paper, we have presented RemoteCLIP, the first general-purpose vision-language foundation model for remote sensing. During developing such a foundation model, our key insights are two-fold. First, CLIP models, which are pretrained on massive image-text pairs collected from the internet, are surprisingly powerful models for remote sensing tasks. Secondly, although in-domain fine-tuning (\textit{i.e.,} continual pretraining) significantly improves the performance, data quantity becomes a major bottleneck in this process, especially when we attempt to specialize large CLIP models in the remote sensing domain.

Based on the above observations, we developed a pipeline for data scaling and subsequently tune the CLIP models on the expanded dataset. The resulting RemoteCLIP model showed superior results on downstream tasks. Importantly, despite simplicity, RemoteCLIP still established a series of SOTA performances on various benchmarks. It highlights the importance of data-centric methodology in developing foundation models. Such observation is also in line with other efforts of building in-domain foundation models, such as BioMedCLIP~\cite{Zhang2023LargeScaleDP} in the medical domain.

\textcolor{black}{Nevertheless, RemoteCLIP still has several known limitations, and we would like to address these issues in our future works:}

\begin{itemize}
    \item \textcolor{black}{Our largest RemoteCLIP model, initialized from OpenAI's ViT-Large-14 CLIP model, boasts 304M parameters in its visual backbone and was trained on 400M data. Despite being significantly larger than previous remote sensing retrieval models, there is ample room for further scaling up. For instance, Billion-scale MAE~\cite{Cha2023ABF} demonstrated that a ViT with a 2B scale can be successfully applied to remote sensing imagery. In the future, we plan to increase the number of model parameters to enhance the capacity of RemoteCLIP models.}
    
    \item \textcolor{black}{\textbf{Data Scaling}: Scaling up the model size necessitates a simultaneous expansion of the data scale. Although the RemoteCLIP data is already 12 times larger than the combination of all adopted image-text data in this paper, it may still be insufficient to train a much larger model. In the future, we aim to expand the pretraining data by incorporating weakly-labeled data (classification datasets) and unlabeled data (via pseudo labeling).}
    
    \item \textcolor{black}{\textbf{Data Quality and Diversity}: The quality and diversity of data are crucial. While our B2C and M2B approach effectively translates heterogeneous annotations (e.g., bounding box and segmentation maps) into homogeneous captions, our rule-based conversion methodology has limited diversity. In our future work, we plan to generate richer captions by introducing generative language models. Additionally, the modality diversity of RemoteCLIP is currently limited, and exploring more sensory modalities beyond RGB is a promising direction.}
    
\end{itemize}

\section*{Acknowledgments}
% This work was partially supported by Joint Fund of Ministry of Education for Equipment Pre-research (8091B022123), Research Fund from Science and Technology on Underwater Vehicle Technology Laboratory (2021JCJQ-SYSJJ-LB06905), Key Laboratory of Information System Requirements, No. LHZZ 2021-M04, Water Science and Technology Project of Jiangsu Province under grant No. 2021063, Qinglan Project of Jiangsu Province.
This work was partially supported by National Nature Science Foundation of China (62372155 and 32371877), Aeronautical Science Fund (2022Z071108001), Joint Fund of Ministry of Education for Equipment Pre-research (8091B022123), Water Science and Technology Project of Jiangsu Province under grant No. 2021063, Technology Winter Olympics Special Project (201001D) Forest Fire Comprehensive System Construction-Unmanned Aerial Patrol Monitoring System of Chongli (DA2-20001), and Qinglan Project of Jiangsu Province.

\footnotesize
\bibliographystyle{IEEEtran}
\bibliography{RemoteCLIP}

\begin{IEEEbiography}[{\includegraphics[width=1in,height=1.25in,clip,keepaspectratio]{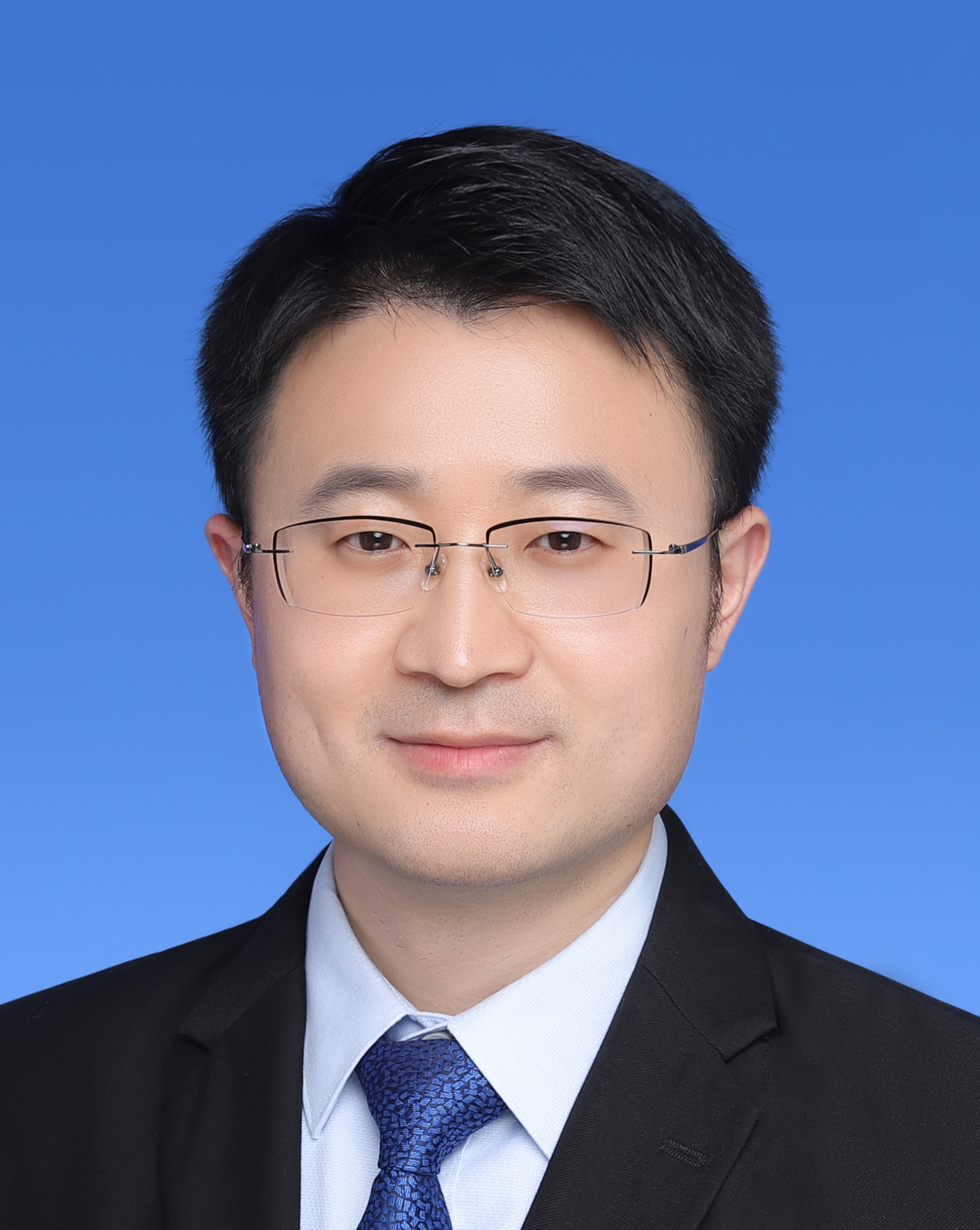}}]{Fan Liu}
(Member, IEEE) is currently a professor at Hohai University. He received his B.S. degree and Ph.D. degree from Nanjing University of Science and Technology (NUST) in 2009 and 2015.  From September 2008 to December 2008, he studied at Ajou University in South Korea. From February 2014 to May 2014, he worked at Microsoft Research Asia. His research interests include computer vision, pattern recognition, and machine learning. Dr. Liu serves as a reviewer of \emph{IEEE TNNLS},  \emph{IEEE TKDE},  \emph{ACM TIST},  \emph{Information Sciences},  \emph{Neurocomputing},  \emph{Pattern Analysis and Application} and an executive director of Jiangsu association of Artificial Intelligence (JSAI).
\end{IEEEbiography}

\begin{IEEEbiography}[{\includegraphics[width=1in,height=1.25in,clip,keepaspectratio]{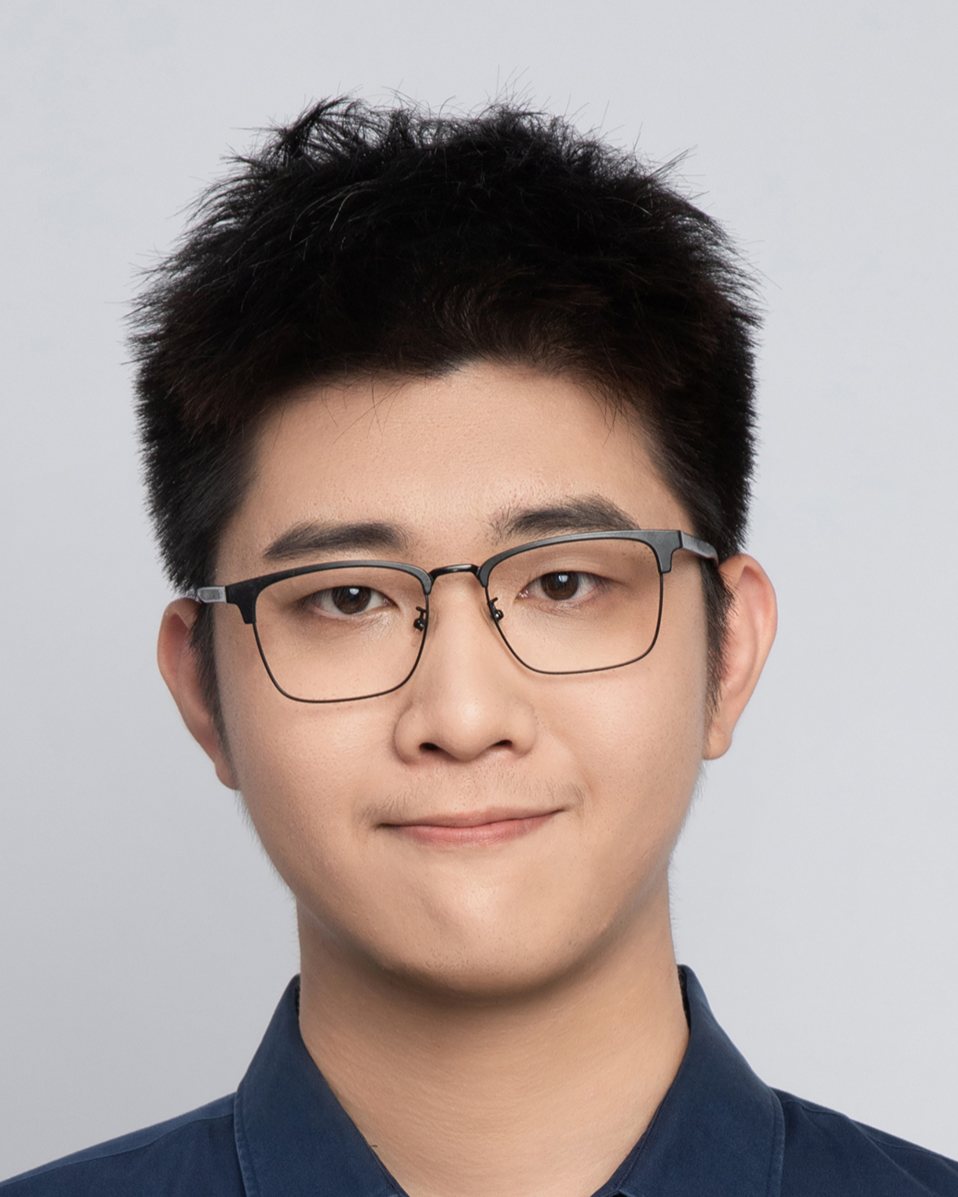}}]{Delong Chen} 
received the B.Eng. degree in computer science from Hohai University, Nanjing, China, in 2021. He is currently pursuing the Ph.D. degree at The Hong Kong University of Science and Technology (HKUST), Hong Kong. He received the Best Demo Award in IEEE ICME’21, the Best Paper Award in AAAI’23 Inaugural Summer Symposium, and the LTDL Best Dataset Paper in IJCAI’21. His research interests include vision-language and representation learning.
\end{IEEEbiography}

\begin{IEEEbiography}[{\includegraphics[width=1in,height=1.25in,clip,keepaspectratio]{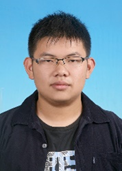}}]{Zhangqingyun Guan} received the B.E. degree in computer science from Changzhou University, Nanjing, China, in 2022. He is a student pursuing the M.S. degree in Hohai University, Nanjing, China. His research interests include image-text retrieval, vision-language learning and multimodal learning.
\end{IEEEbiography}

\begin{IEEEbiography}[{\includegraphics[width=1in,height=1.25in,clip,keepaspectratio]{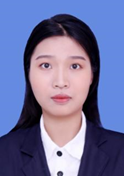}}]{Xiaocong Zhou} received the B.S. degree in information and computing science from Hohai University, Nanjing, China, in 2022. She is currently pursuing the M.S. degree in computer science with Hohai University, Nanjing, China. Her research interests include image captioning, vision-language learning and self-supervised learning.
\end{IEEEbiography}

\begin{IEEEbiography}[{\includegraphics[width=1in,height=1.25in,clip,keepaspectratio]{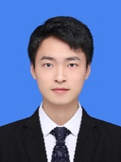}}]{Jiale Zhu} received the B.E. degree in computer science from Hohai University, Nanjing, China, in 2022. He is currently pursuing the M.S. degree in computer science with Hohai University, Nanjing, China. His research interests include semantic segmentation and vision-language learning.
\end{IEEEbiography}

\begin{IEEEbiography}[{\includegraphics[width=1in,height=1.25in,clip,keepaspectratio]{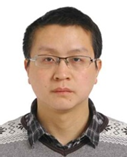}}]{Liyong Fu} received the B.S. degree in forestry from Shanxi Agriculture University, Jinzhong, China, in 2007, the M.S. degree in forest biometrics from Nanjing Forestry University, Nanjing, China, in 2009, and the Ph.D. degree in forest biometrics from the Chinese Academy of Forestry, Beijing, China, in 2012. 
He is currently a Full Professor of forest biometrics with the Department of Forest Management and Statistics, Chinese Academy of Forestry. He has authored or coauthored more than 32 SCI articles in prestigious peer-reviewed international journals, including Briefings in Bioinformatics and Neural Networks, during the recent 5 years. 

Dr Fu was a recipient of the first prize of Liang Xi Best Paper Award for Young Scholars once and the second prize twice and the Fourteenth Young Science and Technology Award of China Forestry in 2017. He was first time selected as one of “Chinese Young Talent” supported by the China Association for Science and Technology in 2016. He is currently an Editorial Board Member for Forestry: An International Journal of Forest Research and a Guest Editor of Remote Sensing Journal. He is currently a senior research engineer at the Nanjing Research Institute of Electronic Engineering. His research interests include intelligent command and control system, deep reinforcement learning, swarm intelligence.
\end{IEEEbiography}

\begin{IEEEbiography}[{\includegraphics[width=1in,height=1.25in,clip,keepaspectratio]{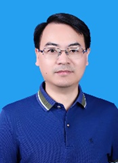}}]{Qiaolin Ye} (Member, IEEE) received the B.S. degree in computer science from the Nanjing Institute of Technology, Nanjing, Jiangsu, China, in 2007, the M.S. degree in computer science and technology from Nanjing Forestry University, Nanjing, in 2009, and the Ph.D. degree in pattern recognition and intelligence systems from the Nanjing University of Science and Technology, Nanjing, in 2013. He is currently an Associate Professor with the Department of Computer Science, Nanjing Forestry University, and the Key Laboratory of Intelligent Information Processing, Nanjing Xiaozhuang University, Nanjing. He has authored over 50 scientific articles. Some of them are published in the \emph{IEEE Transactions on Neural Networks and Learning Systems}, the \emph{IEEE Transactions on Information Forensics and Security}, and the \emph{IEEE Transactions on Circuits and Systems for Video Technology}. His research interests include machine learning, data mining, and pattern recognition.
\end{IEEEbiography}

\begin{IEEEbiography}[{\includegraphics[width=1in,height=1.25in,clip,keepaspectratio]{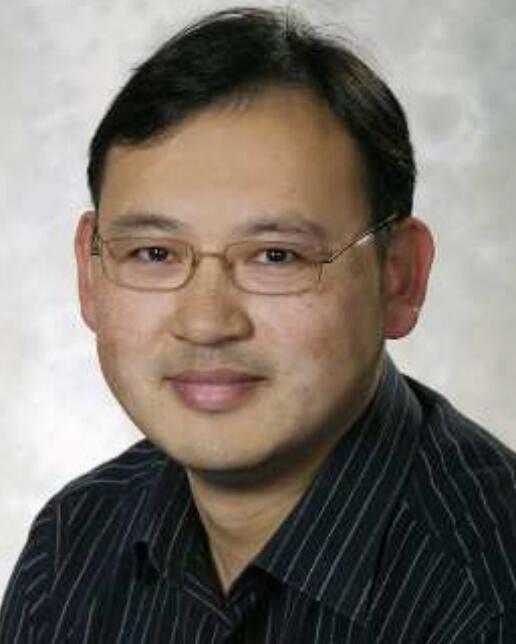}}]{Jun Zhou} (Senior Member, IEEE) received the B.S. degree in computer science and the B.E. degree in international business from the Nanjing University of Science and Technology, Nanjing, China, in 1996 and 1998, respectively, the M.S. degree in computer science from Concordia University, Montreal, QC, Canada, in 2002, and the Ph.D. degree from the University of Alberta, Edmonton, AB, Canada, in 2006.

In June 2012, he joined the School of Information and Communication Technology, Griffith University, Nathan, QLD, Australia, where he is currently a professor. Before this appointment, he was a Research Fellow with the Research School of Computer Science, Australian National University, Canberra, ACT, Australia, and a Researcher with the Canberra Research Laboratory, NICTA, Canberra. His research interests include pattern recognition, computer vision, and spectral imaging with their applications in remote sensing and environmental
informatics. Dr. Zhou is an Associate Editor of \emph{IEEE Transactions on Geoscience and Remote Sensing} and \emph{Pattern Recognition Journal}. 

\end{IEEEbiography}

\end{CJK*}\end{document}